\documentclass[11pt,a4paper]{article}
\usepackage[left = 1in, right = 1in]{geometry}

\usepackage{algorithm}
\usepackage[noend]{algpseudocode}
\usepackage{graphicx}
\usepackage{subfigure}
\usepackage[hyphens]{url}
\usepackage{booktabs} % for professional tables
\newcommand{\Break}{\State \textbf{break} \\}
\usepackage[unicode=true]{hyperref}
\usepackage[numbers]{natbib} 

\usepackage[makeroom]{cancel}
\usepackage[utf8]{inputenc} % allow utf-8 input
\usepackage[T1]{fontenc}    % use 8-bit T1 fonts
\usepackage{hyperref}       % hyperlinks
\usepackage{url}            % simple URL typesetting
\usepackage{booktabs}       % professional-quality tables
\usepackage{amsfonts}       % blackboard math symbols
\usepackage{nicefrac}       % compact symbols for 1/2, etc.
\usepackage{microtype}      % microtypography
\usepackage{lipsum}
\usepackage{graphicx}
\graphicspath{ {./images/} }
\usepackage{natbib} 
\usepackage{amsmath}
\usepackage{amssymb}
\usepackage{mathtools}
\usepackage{amsthm}
\usepackage{bbm}

\usepackage[capitalize,noabbrev]{cleveref}

\theoremstyle{plain}

\newtheorem{theorem}{Theorem}[section]

\newtheorem{corollary}[theorem]{Corollary}
\theoremstyle{definition}

\theoremstyle{remark}

\usepackage[textsize=tiny]{todonotes}

\begin{document}

\title{Conformal Tail Risk Control for Large Language Model Alignment \thanks{We thank Wanning Chen and Wanqiao Xu for helpful comments. L.L. is grateful for the support of National Science Foundation grant DMS-2338464. }}

\author{
 Catherine Yu-Chi Chen \thanks{Institute for Computational and Mathematical Engineering, Stanford University. Email: \texttt{cyc2152@stanford.edu}} \\
  \and Jingyan Shen \thanks{Department of Industrial Engineering and Operations Research, Columbia University. Email: \texttt{js5544@columbia.edu}}
  \and Zhun Deng \thanks{Department of Computer Science, University of North Carolina at Chapel Hill. Email: \texttt{zhundeng@cs.unc.edu}}
  \and Lihua Lei \thanks{Graduate School of Business and Department of Statistics, Stanford University. Email: \texttt{lihualei@stanford.edu}}
}

\maketitle
\begin{abstract}
Recent developments in large language models (LLMs) have led to their widespread usage for various tasks. The prevalence of LLMs in society implores the assurance on the reliability of their performance. In particular, risk-sensitive applications demand meticulous attention to unexpectedly poor outcomes, i.e., tail events, for instance, toxic answers, humiliating language, and offensive outputs. Due to the costly nature of acquiring human annotations, general-purpose scoring models have been created to automate the process of quantifying these tail events. This phenomenon introduces potential human-machine misalignment between the respective scoring mechanisms. In this work, we present a lightweight calibration framework for blackbox models that ensures the alignment of humans and machines with provable guarantees. Our framework provides a rigorous approach to controlling any distortion risk measure that is characterized by a weighted average of quantiles of the loss incurred by the LLM with high confidence. The theoretical foundation of our method relies on the connection between conformal risk control and a traditional family of statistics, i.e., L-statistics. To demonstrate the utility of our framework, we conduct comprehensive experiments that address the issue of human-machine misalignment. 
\end{abstract}

% keywords can be removed

\section{Introduction}
Large Language Models (LLMs) have proven to be pervasive in society with applications across various settings, including those of high sensitivity. While LLMs generally perform quite well, there remains a low probability associated with the event that the models generate undesirable and even catastrophic outputs, including misinformation, malicious use cases, and harmful/toxic comments. Although these responses are relatively rare, each occurrence has the potential to cause significant harm to individuals or even society as a whole.

Quantitative notions of disutility, such as toxicity, are typically based on human annotations that are costly to acquire.  
To reduce labor cost, general purpose models, for example, Detoxify \cite{detoxify}, have been created to automatically generate disutility measures for LLM outputs. Given the existence of both human and machine assessments, a common issue is the misalignment of the two metrics, which might be caused by distribution shift of human opinions when deploying the model. This is further complicated by the lack of an inherent scale that governs machine scores. While many techniques, such as, Reinforcement Learning from Human Feedback (RLHF), have been proposed to improve alignment, these approaches typically do not provide guarantees regarding the alignment of machine-generated scores. Even when such guarantees exist, they often rely on assumptions that are hard to defend. Moreover, many RLHF approaches require computationally intensive model refitting \citep{christiano2017deep, ziegler2019fine,casper2023open}. 

In this work, we address the issue of misalignment through the lens of risk control. {In particular, we treat the human-annotated disutility score as the ground-truth risk measure and calibrate the raw outputs generated by the LLM to control certain functionals of the risk distribution at a pre-specified level. Unlike existing methods, our method does not involve model refitting and provides finite-sample guarantees of risk control under no assumptions about the LLM or the underlying data generation process.}

The proposed framework substantially generalizes the existing literature in conformal risk control \cite{dfrcps,ltt,angelopoulos2023conformalriskcontrol}, which only holds for traditional risk measures characterized by the expectation of a loss function common in supervised learning problems. {These risk measures are not suitable for tail risks associated with low probability events. By contrast, other works only consider risk measures that are quantiles of a loss function on the human scores of the outputs \citep{mohri2024languagemodelsconformalfactuality, cherian2024largelanguagemodelvalidity, quach2024conformallanguagemodeling}. While taking into account the rare events, quantile risk measures often underestimate the tail risk because they do not account for information from the more extreme quantiles. A better suited metric is the Conditional Value-at-Risk (CVaR) \cite{cvar}, which measures the average of a range of upper quantiles. The CVaR is a special case of distortion risk measures, defined as weighted averages of loss quantiles, which is considered in \cite{qrc} and \cite{prc}. In this work, we explore how distortion risk control can be applied to align LLMs with respect to any disutility metric and leverage techniques from the theory of L-estimators in statistics \cite{vandervaart} to achieve finite-sample control of any distortion risk measure.} Our newly derived risk control bounds are also tighter than previous work \cite{qrc}. 

\section{A risk-controlling approach for LLM alignments}

\subsection{Problem setup}
An LLM produces a response $y(x) \in \mathcal{Y}$ for any given user prompt $x \in \mathcal{X}$ from a distribution $p(y \mid x)$. The response $y(x)$ could be an answer to a question or a response to a comment made by the user. To evaluate the disutility of $y(x)$, human-annotators are enlisted to rate different aspects of $y$, for instance, misinformation and toxicity. Let $r(y)$ denote the human rating of $y$, which is generally random due to cognitive uncertainty. Throughout the paper we assume that $r(y) = 0$ when the LLM declines to respond. An example of a disutility measure is shown in the Jigsaw Unintended Bias in Toxicity Classification \cite{jigsaw} dataset, which provides toxicity labels from up to 10 human-annotators for each of the 2 million comments on the Civil Comments platform. The total score $r(y)$ can be obtained by averaging over the ratings of the annotators. For general disutility metrics, $r(y)$ can also be defined as the negative reward estimated using the Bradley-Terry model, as done in RLHF.

In most applications, one can leverage historical data to train a model to guess the human-rated disutility. For example, Detoxify \citep{detoxify} is a machine learning model that assesses the toxicity of responses. Denote ${r}_m(y)$ as the machine-generated disutility score for a response $y$. Note that ${r}_m(y)$ typically differs from $r(y)$ and may even lack monotonicity with respect to $r(y)$. Although machine ratings are inexpensive and scalable, the misalignment, or lack of rank preservation between the machine and human ratings diminishes its reliability. Moreover, it is often hard to interpret the scale of the score and choose the right cut-off to decide whether the generated data should be accepted, especially when applied to different contexts. For example, a sentence with a toxicity score of 0.85 could be very toxic for workplace conversations but acceptable in historian studies. Similarly, the RLHF reward only captures ordinal preferences rather than cardinal values.

Given any generative model or sampler $p(y \mid x)$, an aligned model seeks to produce an output $\tilde{y}(x)$ in a way such that the overall human-rated disutility, $r(\tilde{y}(x))$, is minimized. {We call $\tilde{y}(x)$ a calibrated model.}

Let $x$ be a random draw from the population of prompts of interest, and  $F_{r(\tilde{y}(x))}$ denote the distribution of $r(\tilde{y}(x))$ over $x$, randomness of $\tilde{y}(\cdot)$, and cognitive uncertainty of $r(\cdot)$. 
To aggregate over different prompts and integrate out the randomness of responses and cognitive uncertainty, we can define a summary measure $R \left(F_{r(\tilde{y}(x))} \right )$ where $F_{r(\tilde{y}(x))}$ denotes the cumulative distribution function (CDF) of $r(\tilde{y}(x))$ with $x$ being a draw from a population of prompts, and $R(\cdot)$ being a functional that maps any distribution to a non-negative number. For example, $R(F)$ can be chosen as the mean disutility. 

To achieve alignment, we aim to control the risk $R \left(F_{r(\tilde{y}(x))} \right )$ at a pre-specified level $\alpha$. This objective is very different from existing practices (e.g., RLHF) that target the risk associated with the machine disutility rather than the human disutility scores. It is a challenging task because the human rating function $r(\cdot)$ operates as a black box, and researchers can only observe its output for a given set of prompt-response pairs. 

Note that any risk level $\alpha$ can be achieved by abstaining from responding to any prompts. Clearly, this should be avoided. As will be seen later, the deployment cost of our calibrated LLM increases as $\alpha$ decreases and could grow to infinity as $\alpha \rightarrow 0$, if no abstention is allowed. As a result, though being conservative may appear innocuous, it is unnecessarily costly. To minimize deployment costs, we would want $R \left(F_{r(\tilde{y}(x))} \right )$ to be as close to $\alpha$ as possible.

\subsection{Choice of risk functions}

In classical statistical decision theory, risk is often defined as the expectation of a loss function, i.e. $R(F) = \mathbb{E}_{r \sim F}[L(r)]$, where $L(\cdot)$ is a loss function. However, considering that the majority of generated data is normal, traditional risk measures may fail to capture the disutility as these events are only manifested in the tail. 

\begin{figure}
    \centering
    \includegraphics[width=0.6\linewidth]{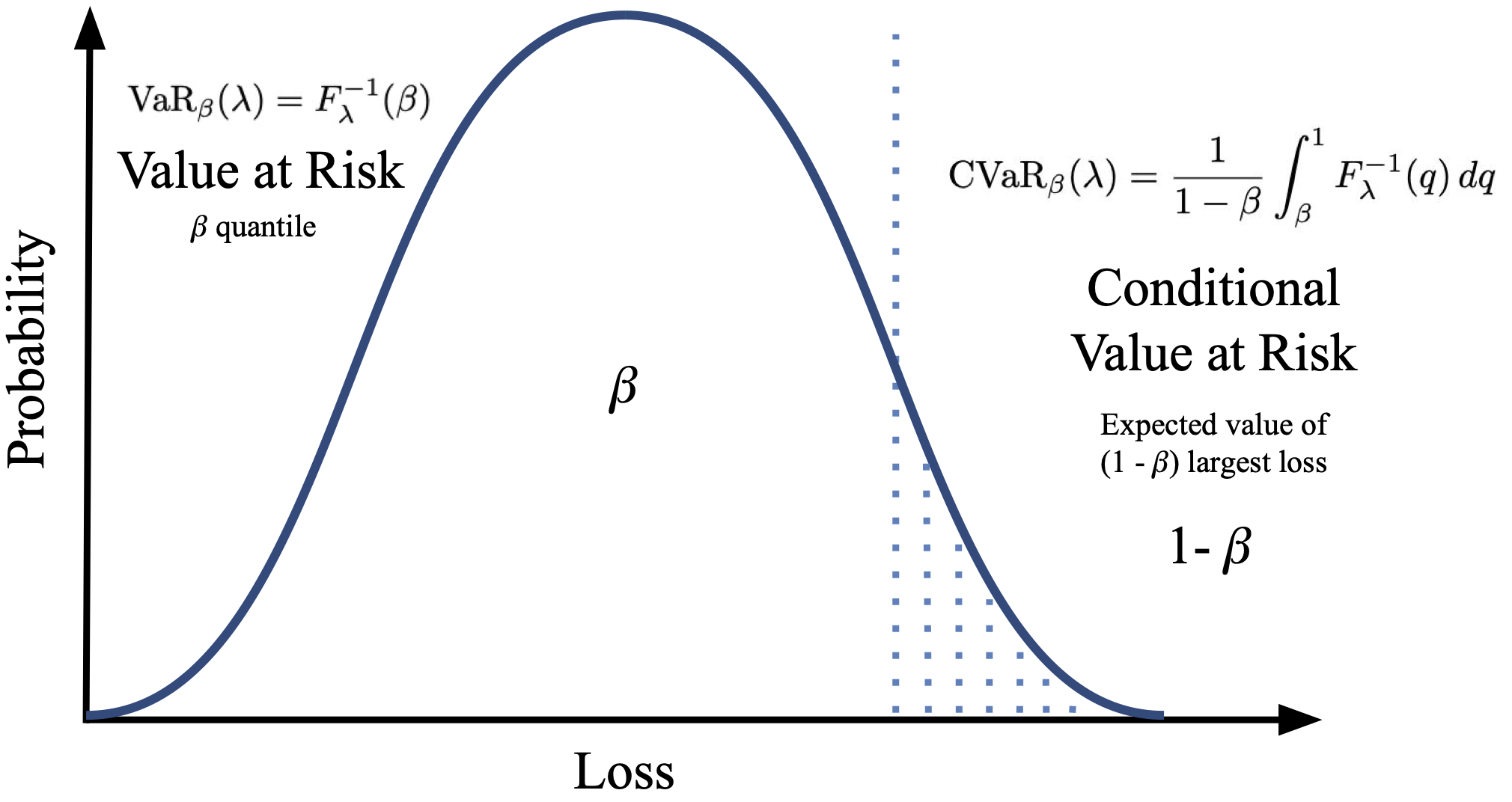}
    \caption{\textbf{Examples of distortion risk measures: Value-at-Risk (VaR) and Conditional Value-at-Risk (CVaR).}}
    \label{fig:distortion}
\end{figure}

Instead, we choose $R(F)$ to be a \textit{distortion risk measure} \citep[e.g.][]{balbas2009properties,qrc}, defined as a weighted average of quantiles,
\begin{equation}\label{eq:distortion_risk}
R_{\psi}(F):= \int_0^1  F^{-1}(p) \, d \psi
\end{equation}
where $F^{-1}(p)\stackrel{\Delta}{=} \operatorname{inf}\{x: F(x) \geq p\}$ denotes the $p$-th quantile of $F$, and $\psi(\cdot)$ is a weighting measure such that $\psi(p) \geq 0$ and $\int_0^1 d\psi(p) = 1$. An example of a distortion risk measure is the widely-used $\operatorname{CVaR}_{\beta}$ \cite{cvar} where $\psi$ is the uniform measure on $[\beta, 1]$. Other examples include mean (with $\psi$ the uniform measure on $[0,1]$) and $\beta$-th quantile, also known as Value-at-Risk, for any $\beta$ (with $\psi$ the point mass at $p$). Examples of distortion risk measures are shown in (Fig. \ref{fig:distortion}). We also want to remark that expected mean and quantiles are also special cases of distortion risk measures.

\subsection{Augmentation of LLM outputs}\label{subsec:augmentation}
If the original output $y(x)$ does not control the risk at $\alpha$, we need to refine it. Instead of modifying the underlying data generating process $p(y \mid x)$ based on the score by retraining the model, we adopt a light-weight calibration approach that amounts to tuning a one-dimensional parameter instead of the high-dimensional model parameters. Specifically, for every prompt $x$, we request a response from the LLM multiple times to generate a candidate set $\mathcal{C}(x) = \{y_1(x). \ldots, y_N(x)\}$, where $N$ denotes the set size. To maximize the information content, we follow \cite{quach2024conformallanguagemodeling} in eliminating responses to ensure diversity, quality, and set-confidence; see Appendix \ref{subsec:conformal_language_model} for details. 

Our idea is to generate a nested sequence of subsets of $\mathcal{C}(x)$ that have increasing disutility. Since we are not allowed to collect human ratings on the fly, we use the machine disutility score as a proxy. In particular, for each $\lambda$ in the range of machine scores $\Lambda$, we generate a set $$\mathcal{C}_\lambda(x) = \{y\in \mathcal{C}(x): {r}_m(y) < \lambda\}$$
by using the machine disutility score $r_m(\cdot)$. Without loss of generality, we assume $\Lambda = [\lambda_{\min}, \lambda_{\max} ]$. We assign a disutility score $r_{\lambda}(x)$ for $C_\lambda(x)$ as the worst human rating, i.e.,
\[r_{\lambda}(x) = \max_{y\in \mathcal{C}_\lambda(x)}r(y).\]
By design, $r_{\lambda}(x)$ is non-decreasing in $\lambda$, a key property that our method leverages. Moreover, $\mathcal{C}_{\lambda_{\min}}(x) = \emptyset$ for any $x$ and hence $r_{\lambda_{\min}}(x) = 0$. The process of generating $\mathcal{C}_\lambda(x)$ and $r_\lambda(x)$ is illustrated in Figure \ref{fig:procedure}.

 Notably, $\mathcal{C}_\lambda(x)$ can be computed for any $x$ and $\lambda$, because it relies solely on machine disutility scores, whereas $r_\lambda(x)$ is a black-box function that is only available for prompts with collected human ratings. Our goal is to choose $\hat{\lambda}$ based on human-annotated data such that $R(F_{r_{\hat{\lambda}}(x)})\le \alpha$. Since $r_{\hat{\lambda}}(x)$ is defined as the worst human score in $\mathcal{C}_{\hat{\lambda}}(x)$, we can pick any candidate $\tilde{y}(x) \in \mathcal{C}_{\hat{\lambda}}(x)$ and the resulting risk $R(F_{r(\tilde{y}(x))})$ will be controlled at level $\alpha$. When $\mathcal{C}_{\hat{\lambda}}(x)$ is empty, the calibrated LLM simply declines to respond. Importantly, the selection can be arbitrary -- for example, we could choose the response from $\mathcal{C}_{\hat{\lambda}}(x)$ that has the minimal machine disutility score or maximal information content measured by another metric. To summarize, we reduce the task from retraining a calibrated LLM $\tilde{y}(x)$ with high-dimensional model parameters to searching for a univariate parameter $\hat{\lambda}$. 

\section{Distortion risk control via L-statistics}\label{sec:DRC}

\subsection{Theoretical setting}\label{subsec:setting}
Suppose we sample $n$ prompts $x_1, \ldots, x_n$ i.i.d. from a distribution. For each prompt, we  generate the candidate set $\mathcal{C}(x_i)$ as described in Section \ref{subsec:augmentation}, then recruit human raters to score all responses in the set. Following the procedure in Section \ref{subsec:augmentation}, we can obtain a dataset $\mathcal{D}$ that includes $(x_i, \{r_{\lambda}(x_i)): \lambda\in \Lambda\})$ for $i = 1, \ldots, n$. We assume that the machine disutility score model is pretrained and independent of our dataset $\mathcal{D}$. Then the data points $(x_i, \{r_{\lambda}(x_i)): \lambda\in \Lambda\})$ remain i.i.d. Throughout the rest of the section we denote a generic draw from the prompt distribution by $x$.

For each $\lambda\in \Lambda$, we use the shorthand notation $R_{\psi}(\lambda)$ for  $R_{\psi}(F_{r_\lambda(x)})$, where $R_{\psi}$ is the distortion risk measure defined in \eqref{eq:distortion_risk} with a user-chosen weight measure $\psi$. Our goal is to learn $\hat{\lambda}$ from $\mathcal{D}$ such that 
\[\mathbb{P}_{\mathcal{D}}(R_{\psi}(\hat{\lambda})\le \alpha)\ge 1 - \delta,\]
for some pre-specified $(\alpha, \delta)$. Above, $\mathbb{P}_{\mathcal{D}}$ accounts for the randomness in $\mathcal{D}$. The parameter $\alpha$ represents the target risk level and $1-\delta$ corresponds to the confidence level.  This formulation aligns with the standard objective of probably approximately correct (PAC) learning and has also been studied in the context of conformal risk control \citep{dfrcps, ltt}.

\subsection{L-statistics}

L-statistics refers to the class of estimators expressed as linear combinations of order statistics originating from \cite{mosteller1946some}. Notable examples include the sample quantile, the trimmed mean, and the winsorized mean \citep{tukey1962future}. Fixing $\lambda\in \Lambda$, let $r_{\lambda, (1)} \le r_{\lambda, (2)}\le \ldots \le r_{\lambda, (n)}$ denote the ordered statistics of $(r_\lambda(x_1), \ldots, r_\lambda(x_n))$. Furthermore, let $F_\lambda$ and $\hat{F}_{n, \lambda}$ be the true and empirical distributions of $(r_\lambda(x_1), \ldots, r_\lambda(x_n))$, respectively. For a given distortion risk measure, the plug-in estimator that replaces the true distribution $F$ by the empirical distribution $\hat{F}_n$ is 
\begin{equation}\label{eq:L_stat_dr}
\hat{R}_\psi(\lambda) = R_\psi(\hat{F}_{n,\lambda}) = \int \hat{F}_{n,\lambda}^{-1}(p)d\psi(p).
\end{equation}
By definition, $\hat{F}^{-1}_{n, \lambda}(p) = r_{\lambda, (i)}$ for any $p \in \left(\frac{i-1}{n}, \frac{i}{n}\right]$, thus it can be written as an L-statistic
\[\hat{R}_\psi(\lambda) = \sum_{i=1}^{n}\left\{\psi\left(\frac{i}{n}\right) - \psi\left(\frac{i-1}{n}\right)\right\} r_{\lambda, (i)}.\]

Theorem 22.3 of \cite{vandervaart} describes the asymptotic normality of L-statistics.
Given this result, we show that $\hat{R}_\psi(\lambda)$ is an asymptotically normal estimator of $R_\psi(\lambda)$ for any fixed $\lambda\in \Lambda$
{and we can find a consistent variance estimator. The proof of Theorem \ref{thm:drc} is presented in Appendix \ref{sec:proofs}.}

{
\begin{theorem}
\label{thm:drc}
Assume $r_{\lambda}(x)\in [a, b]$ almost surely for some $-\infty < a < b < \infty$, $F_\lambda$ is continuous and strictly increasing. Further, assume that $\psi(y) = \int_{0}^{y}\psi^{\prime}(z) dz$ for some $\psi'$ that is bounded and continuous at $F_{\lambda}(r)$ for Lebesgue almost-every $r$. Then, 
$$\frac{{\sqrt{n}}(\hat{R}_{\psi}(\lambda) - R_{\psi}(\lambda))}{\hat{\sigma}(\lambda)}\stackrel{d}{\rightarrow}N(0, 1),$$
where  
\begin{equation*}
\hat{\sigma}^2(\lambda) = \int\int {\psi}^{\prime}(\hat{F}_{n, \lambda}(r)) {\psi}^{\prime}(
\hat{F}_{n, \lambda}(\tilde{r}))D_{\lambda}(r, \tilde{r})\, dr \, d\tilde{r},
\end{equation*}
with $\hat{F}_{n,\lambda}$ being the empirical distribution of $r_\lambda(x_1), \ldots, r_\lambda(x_n)$ and
$$D_{\lambda}(r, \tilde{r}) = \hat{F}_{n,\lambda}(r \wedge \tilde{r}) - \hat{F}_{n,\lambda}(r)\hat{F}_{n,\lambda}(\tilde{r}).$$
Equivalently, 
\begin{equation}\label{eq:hatsigma}
\hat{\sigma}^2(\lambda) = \frac{1}{n^2}\sum_{i,j=1}^{n} \psi'\left(\frac{i}{n}\right)\psi'\left(\frac{j}{n}\right)\left(\frac{i\wedge j}{n} - \frac{ij}{n^2}\right).
\end{equation}
\end{theorem}
We assume the boundedness of $r_\lambda(x_i)$ for simplicity. It is a reasonable assumption for ratings which are often designed to be bounded. It can be relaxed with more involved proof techniques \citep{gardiner1979asymptotic, stigler1974linear}.

Among all distortion risk measures, $\operatorname{CVaR}_{\beta}$ is the most interpretable and widely-used metric. For $\operatorname{CVaR}_{\beta}$, we can find a much simpler variance estimator. 
\begin{corollary}\label{cor:CVaR_var}
For $\operatorname{CVaR}_{\beta}$ with $\psi(p) = \max\{p - \beta, 0\} / (1 - \beta)$, Theorem \ref{thm:drc} holds with
$$\hat{\sigma}^2(\lambda) =  \frac{1}{(1-\beta)^2}\widehat{\operatorname{Var}}\left( \{\max\{r_{\lambda}(x_i), r_{\lambda, (\lceil n\beta\rceil)}\}_{i=1}^{n}\}\right),$$
where $\widehat{\operatorname{Var}}$ denotes the sample variance.
\end{corollary}

Another important example is $\operatorname{VaR}_{\beta}$. While the $\operatorname{VaR}$ is not a distortion risk measure with a differentiable $\psi$, we develop parallel theory in Appendix \ref{sec:var} based on the asymptotic theory of empirical quantiles. Unlike $\operatorname{CVaR}_{\beta}$, the asymptotic variance depends on the density $F'_\lambda$ of $r_i$ and hence harder to estimate. We apply the bootstrap technique \citep{efron1994introduction} instead to estimate $\hat{\sigma}^2(\lambda)$. 
}

\begin{figure}
    \centering
    \includegraphics[width=0.6\linewidth]{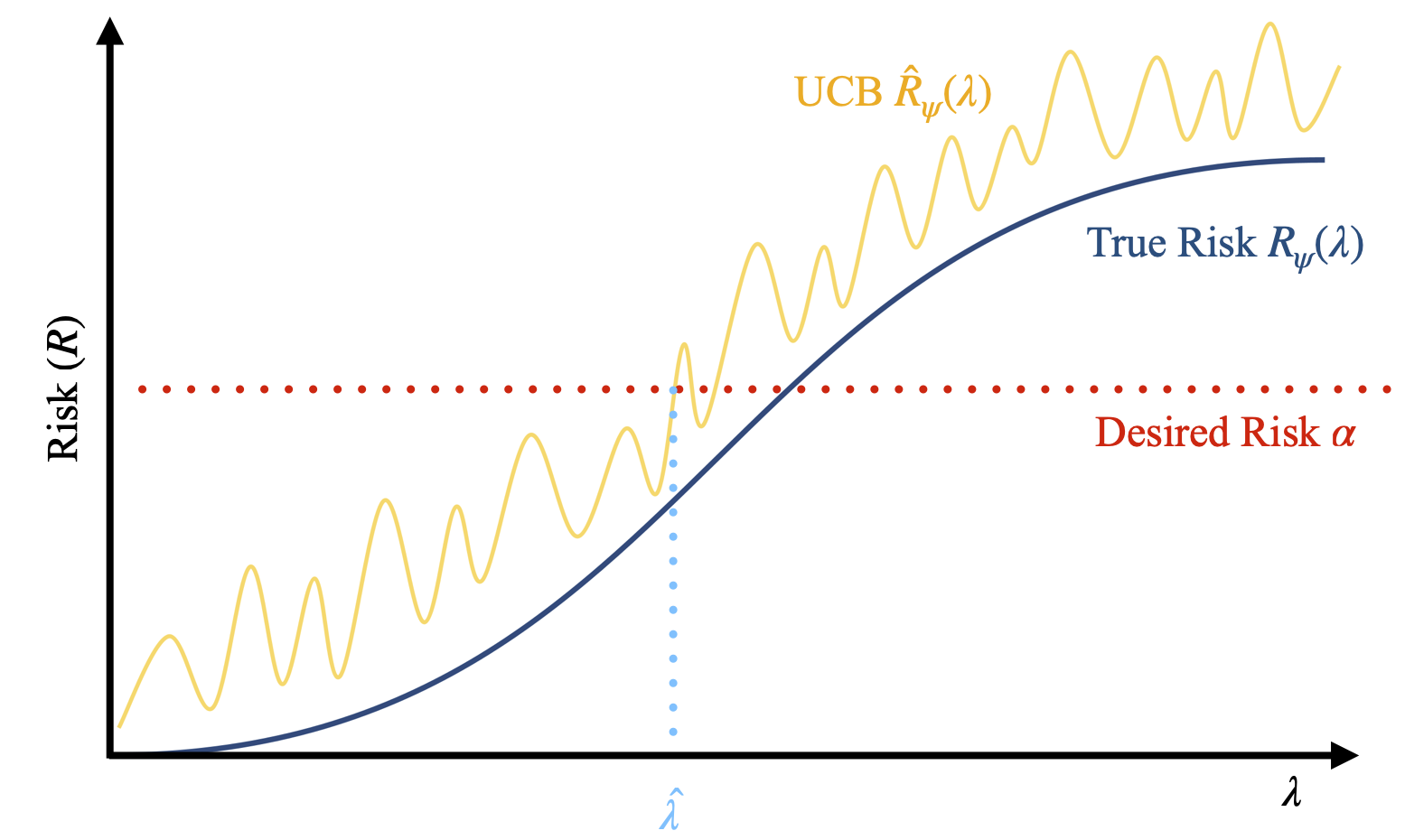}
    \caption{\textbf{Illustration of $\hat{\lambda}$}. We choose $\hat{\lambda}$ as the last $\lambda$ such that the (asymptotic) upper confidence bound $\hat{R}_{\psi}^+(\lambda)$ falls below $\alpha$.} 
    \label{fig:risk-ucb}
\end{figure}

\subsection{Conformal distortion risk control via L-statistics}
\begin{figure*}
    \centering
    \includegraphics[width=0.9\linewidth]{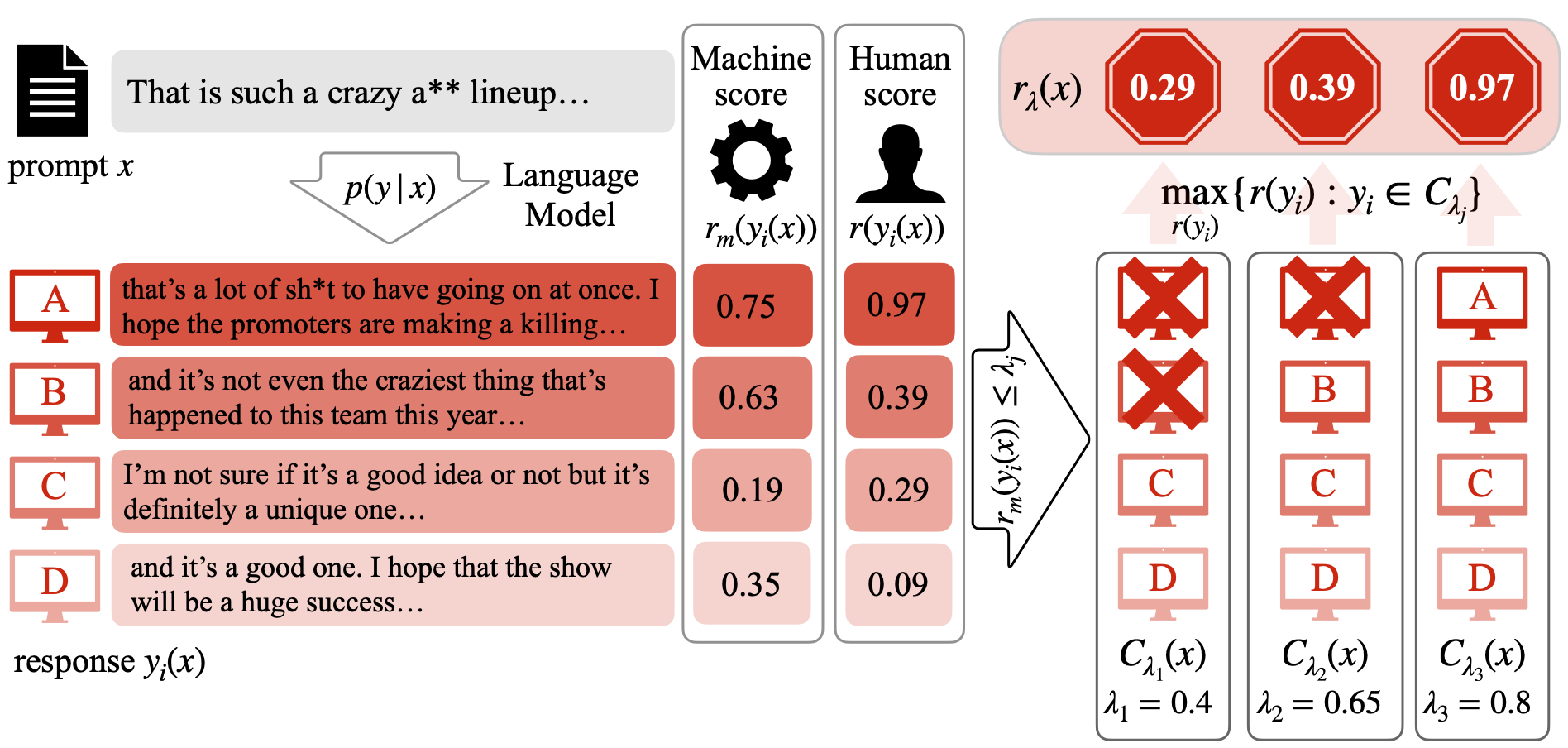}
    \caption{\textbf{Illustration of the process to generate $\mathcal{C}(x), \mathcal{C}_\lambda(x)$, and $r_{\lambda}(x)$}. We sample $N$ responses $y_i$, $i = 1, \ldots, N$ from the LLM $p(y \mid x)$. Each response is associated with a machine disutility score ${r}_m(y_i(x))$, and a human-rated disutility score $r(y_i(x))$. To construct $\mathcal{C}_{\lambda_j}(x)$, we keep the responses such that the machine toxicity score satisfies ${r}_m(y_i(x)) \leq \lambda_j$ for each $\lambda_j \in \Lambda$. Finally, we compute the induced score $r_{\lambda}(x)$ by taking the maximum human disutility score of each $\mathcal{C}_{\lambda_j}(x)$.}
    \label{fig:procedure}
\end{figure*}

By design, $R_{\psi} (\lambda)$ is monotonic in $\lambda$. This allows us to apply the method of \cite{dfrcps} to achieve risk control by inverting a pointwise upper confidence bound (UCB). Specifically, we choose
\begin{equation}\label{eq:lambdahat}
\hat{\lambda} = \max \{ \lambda \in \Lambda :  \hat{R}_{\psi}^+(\lambda')\leq \alpha, \,\forall \lambda' \le \lambda\},
\end{equation}
where 
\[\hat{R}_{\psi}^+(\lambda) = \hat{R}_{\psi}(\lambda) + z_{1-\delta} \cdot \hat{\sigma}(\lambda),\]
and $z_{1-\delta}$ is the $(1-\delta)$-th quantile of the standard normal distribution, as illustrated in Fig. \ref{fig:risk-ucb}. {In practice, we discretize $\Lambda$ to avoid checking the condition $\hat{R}_\psi^+(\lambda)\le \alpha$ for infinitely many points. } Our procedure is outlined in Algorithm \ref{alg:DRC} and illustrated in Fig. \ref{fig:procedure}.

\begin{algorithm}
    \caption{Conformal distortion risk control}
    \label{alg:DRC}
    \begin{algorithmic}
        \Require machine-scoring model $r_m(\cdot)$, discrete subset of its range $\Lambda$, human-annotated scores $r(\cdot)$, set of prompts $\mathcal{X} = (x_i)_{i=1}^{n}$, target level $\alpha$, tolerance level $\delta$, weighting function $\psi$.
        \Function{DRC}{$(\mathcal{X}, \Lambda, r_m(\cdot), r(\cdot), \alpha, \delta, \psi)$}
        \For{$x_i \in \mathcal{X}$}
   \State{$\mathcal{C} \gets \textsc{CandidateSet}(x_i)$} \Comment{See Algorithm \ref{alg:conformal-sampling-with-rejection}}    
   \For{$\lambda \in \Lambda$}

    \State $\mathcal{C}_{\lambda}(x_i) = \{\}$
    \For{$y_k \in \mathcal{C}(x_i)$}
    \If{$r_m(y_k) < \lambda$}
    \State{$\mathcal{C}_{\lambda}(x_i) \gets \mathcal{C}_{\lambda}(x_i) \cup \{y_k\}$}
    \EndIf
    \State{$r_{\lambda}(x_i) \gets \max\{r(y_j): y_j \in \mathcal{C}_{\lambda}(x_i)\}$}
    \EndFor
    \EndFor
    \EndFor

   \For{$\lambda \in \Lambda$}    
    \State{$(r_{\lambda, (1)},\ldots, r_{\lambda, (n)}) \gets \textsc{Sort}(r_\lambda(x_1), \ldots, r_\lambda(x_n))$}
    \State{$\hat{R}_{\psi}(\lambda) \gets \sum_{i=1}^{n} \left\{\psi\left(\frac{i}{n}\right) - \psi\left(\frac{i-1}{n}\right)\right\} r_{\lambda, (i)}$}
    \State{$\hat{\sigma}^2(\lambda)\gets $ Eq.  \eqref{eq:hatsigma}}
    \State{$\hat{R}^{+}_{\psi}(\lambda)\gets \hat{R}_\psi(\lambda) + z_{1-\delta} \cdot \frac{\hat{\sigma}(\lambda)}{\sqrt{n}}$} \Comment{UCB}
    \EndFor
    \State{$\hat{\lambda} \gets \max\left\{ \lambda \in \Lambda : \hat{R}^{+}_{\psi}(\lambda')\leq \alpha, \,\forall \lambda' \le \lambda \right\}$}

   \Return{$\hat{\lambda}$}
   \EndFunction
    \end{algorithmic}
\end{algorithm}

By Theorem \ref{thm:drc}, $\hat{R}_{\psi}^+(\lambda)$ is an asymptotic $(1-\delta)$ UCB. Although \cite{dfrcps} focus on traditional risk measures that are expressed as expected loss, we can easily extend  their result (Theorem 6) to distortion risk measures.

\begin{theorem}\label{thm:drc_rcps}
    Let $\hat{\lambda}$ be defined in \eqref{eq:lambdahat}. {Assume $R_\psi(\lambda)$ is continuous and strictly increasing.} In the same setting as Theorem \ref{thm:drc}
    \begin{equation}\label{eq:exactness}
    \liminf_{n \rightarrow \infty} \mathbb{P}_{\mathcal{D}}\left( R_{\psi} \left( \hat{\lambda}\right) \leq \alpha \right) \ge 1 - \delta.
    \end{equation}
    {As a consequence, for any selection mechanism that picks $\tilde{y}(x)$ from $\mathcal{C}_{\hat{\lambda}}(x)$, 
    \[\liminf_{n \rightarrow \infty} \mathbb{P}_{\mathcal{D}}\left( R_{\psi} \left( F_{r(\tilde{y}(x))}\right) \leq \alpha \right) \ge 1 - \delta.\]}
\end{theorem}

{Remarkably, this result does not involve any assumption on the underlying LLM $p(y\mid x)$ or machine disutility score model $r_m(x)$. In particular, it allows the machine ratings to be arbitrarily misaligned with human ratings.}
  
%\zhun{remove section numbr for appendix, in case there is a change we made later on}

{
\subsection{Alternative (conservative) approaches for distortion risk control}\label{subsec:DKW_BJ}
Another strategy to construct pointwise UCBs for $R_\psi(\lambda)$ is to replace all quantiles by their confidence envelopes. Specifically, if we have statistics $\{\hat{q}_p^+(\lambda): p\in [0,1]\}$ such that 
\begin{equation}\label{eq:upper_confidence_envelope}
\mathbb{P}(q_{p}(\lambda)\le \hat{q}^+_p(\lambda), \,\, \forall p\in [0, 1])\ge 1 - \delta,
\end{equation}
then $\int_{0}^{1}\hat{q}^+_p(\lambda)d \psi(p)$ is a valid $(1-\delta)$ UCB for $R_\psi(\lambda)$. Following \cite{qrc}, we consider two confidence envelopes based on the Dvoretzky-Kiefer-Wolfowitz (DKW) inequality and Berk-Jones (BJ) statistics, respectively. For simplicity, we assume $F_\lambda$ is continuous. 

\paragraph{DKW inequality.} The DKW inequality implies that, for any $\lambda\in \Lambda$ and $\epsilon > 0$,
$$\mathbb{P}\left( \sup_{r\in \mathbb{R}} \mid \hat{{F}}_{n, \lambda}(r) - F_{\lambda}(r) \mid \geq \epsilon \right) \leq 2e^{-2n\epsilon^2}.$$
Letting $\epsilon_{n, \delta} = \sqrt{\log(2/\delta)/2n}$ and $r = q_p(\lambda)$ for $p \in [0, 1]$, we obtain that, with probability at least $1 - \delta$,
\[|\hat{{F}}_{n, \lambda}(q_p(\lambda)) - p|\le \epsilon_{n, \delta}, \quad \forall p\in [0,1].\]
This implies an upper confidence envelope as $\hat{q}_p^+(\lambda) = r_{\lambda, (k_{n, p})}$, 
where $k_{n, p} = \lceil n(p + \epsilon_{n, \delta})\rceil$. Here, if $k_{n, p} > n$, we set $r_{\lambda, (k_{n, p})} = \lambda_{\max}$, the upper bound of $\Lambda$.

\paragraph{Berk-Jones Statistics.} As pointed out by \cite{qrc} and \cite{bates2023testing}, the DKW inequality is overly conservative for $p$ close to $0$ and $1$. The BJ statistic uses the fact that $F_r(\lambda_{\lambda, (i)})\sim \mathrm{Beta}(i, n-i+1)$, where $\mathrm{Beta}$ denotes a Beta-distribution. Let $B_{i, n-i+1}$ denote the cumulative distribution function of $\mathrm{Beta}(i, n-i+1)$. The BJ statistic is then defined as 
\[M_n^{+} = \max_{1\le i \le n}G_{i, n-i+1}(
F_r(r_{\lambda, (i)})).\]
Let $s_{\delta}$ be the $\delta$-th quantile of $M_n^{+}$. Clearly, $s_{\delta}$ does not depend on $F_r$ because $F_r(r_{\lambda, i})\sim \mathrm{Unif}([0,1])$ when $F_r$ is continuous. It can be computed in polynomial time \cite{moscovich2017fast}. Let $s_{ i} = G_{i, n-i+1}^{-1}(s_\delta)$. Then 
\[\mathbb{P}(q_{s_{i}}(\lambda)\le r_{\lambda, (i)}, \,\, \forall i\le n)\ge 1 - \delta.\]
This yields an upper confidence envelope $\hat{q}_{p}^{+}(\lambda) = r_{\lambda, (i)}$ for any $p\in (s_{i-1}, s_{i}]$. For $p > s_n$, we set $\hat{q}_p^{+}(\lambda) = \lambda_{\max}$.

While both DKW and BJ approaches yield finite-sample valid UCBs for $R_\psi(\lambda)$, they are conservative because they do not target the specific choice of $\psi$. In fact, \eqref{eq:upper_confidence_envelope} implies that $\int_{0}^{1}\hat{q}_p^+(\lambda)d\psi(p)$ is a uniform UCB across all weight functions, i.e., 
\[\mathbb{P}\left(R_\psi(\lambda)\le \int_{0}^{1}\hat{q}_p^+(\lambda)d\psi(p), \,\, \forall \psi\right)\ge 1 - \delta.\]
Therefore, the actual coverage (i.e., probability that $R_\psi(\lambda)$ is less than or equal to the above UCB) is typically much higher than $1-\delta$. 
By contrast, the UCB given by the L-statistic is tailored to $R_\psi(\lambda)$ and the coverage converges to $1-\delta$ as $n\rightarrow \infty$. 
}

{
\subsection{Deployment of the calibrated model}

Recall that any choice of $\tilde{y}(x)\in \mathcal{C}_{\hat{\lambda}}(x)$ controls the risk $R_\psi(F_{\tilde{y}(x)})$. For a new prompt $x$, the most cost-effective approach is to sample candidate responses $y_1, y_2, \ldots$ from the underlying LLM $p(y\mid x)$ until the first time the machine disutility score is below $\hat{\lambda}$. Suppose each sample incurs a unit of computational cost. Given the prompt $x$, the sampling cost follows a geometric distribution with rate $\mathbb{P}({r}_m(y) <  \hat{\lambda}\mid x, \hat{\lambda})$. Thus, the (unconditionally) expected cost is 
\begin{equation*}
\mathrm{Cost}(\hat{\lambda}) = \mathbb{E}_{x}\left[\frac{1}{\mathbb{P}({r}_m(y) <  \hat{\lambda}\mid x, \hat{\lambda})}\right].
\end{equation*}
Suppose we have a hold-out set of prompts $\mathcal{D}'$ that is independent of the dataset $\mathcal{D}$ defined in Section \ref{subsec:setting}, $\mathrm{Cost}(\hat{\lambda})$ can be estimated by 
\begin{equation}\label{eq:cost_estimate}
\frac{1}{|\mathcal{D}'|}\sum_{x\in \mathcal{D}'} \frac{1}{\widehat{\mathbb{P}}({r}_m(y) <  \hat{\lambda}\mid x, \hat{\lambda})},
\end{equation}
where the probability is estimated by the Monte-Carlo method.
}

\section{Experiments}

In this section, we perform experiments to investigate the issue of human-machine misalignment by implementing our conformal distortion risk control method to mitigate toxicity of LLM-generated reseponses. This is a critical application, as toxic outputs may cause severely negative impacts on impressionable populations, moreover, propagate across wide audiences, leading to misinformation and harm. 

\subsection{Experimental setup}

\paragraph{Datasets and models} \label{sec:dataset}We randomly draw 10K prompts from the \textsc{RealToxicityPrompts} dataset \cite{gehman2020realtoxicityprompts} . For each selected prompt $x_i$, we generate 40 responses $y_j(x_i)$ using the LLaMA 2 7B model \cite{llama2}. Given the initial responses, we apply the sequential algorithm described in Algorithm \ref{alg:conformal-sampling-with-rejection} to construct the candidate response sets $\mathcal{C}(x_i)$, ensuring the quality of the selected responses. Specifically, we use perplexity (PPL) to evaluate response quality, \textsc{ROUGE-L} to assess similarity between responses, and restrict the maximum set size to 32 as a stopping criterion. More details can be found in Sec. \ref{sec:candidateresponse} of the Appendix.

\paragraph{Toxicity scores} \label{sec:toxicity} 
{To apply our method, we need a human toxicity score function $r(\cdot)$ and a machine toxicity score function $r_m(\cdot)$. Human-annotated data can be costly
and time-consuming to acquire. To evaluate our method, } we create a cheap semi-synthetic benchmark using an existing machine scoring model as the ``human annotator," and a biased model as the ``machine assessor." Specifically, we use  the Detoxify model \cite{detoxify} for $r(\cdot)$ and retrain the Detoxify model for $r_m(\cdot)$ 
 on a biased subset of the Jigsaw Unintended Bias in Toxicity Classification dataset \cite{jigsaw} that consists of $c\%$ of most and least toxic instances. {The goal is to design $r_m(\cdot)$ with varying degree of misalignment from $r(\cdot)$. This allows us to study the effect of misalignment on $\hat{\lambda}$ and hence the cost of tail risk control. In particular, we quantify the misalignment between human and machine scoring models by the Spearman correlation coefficient $\rho$ between the scores across all candidate responses. In our experiment, we train three models for $r_m(\cdot)$ with $c\%\in \{15 \%, 30\%, 70\%\}$. The Spearman correlation coefficients are $0.57, 0.68, 0.78$. More details about these models can be found in Appendix \ref{sec:semisyn}.} 

 \paragraph{Choices of parameters} \label{sec:methods-baselines}
{We consider both $\operatorname{CVaR}_\beta$ and $\operatorname{VaR}_\beta$ control with $\beta \in \{0.5, 0.75, 0.9\}$. We fix the confidence parameter $1-\delta = 0.95$. To determine a reasonable target level $\alpha$, we compute the empirical $\operatorname{CVaR}_q$ on human scores of all candidate responses with $q\in \{1\%, 5\%, 10\%, 15\%, 20\%\}$; see Appendix \ref{sec:alpha}. This suggests a range of reasonable target levels. In particular, we consider $\alpha \in \{0.15, 0.2, 0.25, 0.3, 0.35\}$. }

\paragraph{Evaluation} We randomly split the prompts, using $60\%$ to determine the optimal threshold $\hat{\lambda}$ and the remaining $40\%$ as a held-out test dataset. For each method, after selecting $\hat{\lambda}$, we deploy the calibrated model on the held-out dataset. We then evaluate the realized  $\operatorname{CVaR}_\beta$ and $\operatorname{VaR}_\beta$ of human scores and estimate the sampling cost following \eqref{eq:cost_estimate}. We apply all three versions of conformal distortion risk control based on L-statistics, DKW inequality, and BJ statistics. We refer to them as CDRC-L, CDRC-DKW, and CDRC-BJ, respectively, where CDRC stands for conformal distortion risk control. For CDRC-L and CDRC-DKW, we repeat for $15$ times, and for CDRC-BJ, we repeat for $3$ times due to computational complexity to compute $s_\delta$. 

\subsection{Results}
\begin{figure*}
    \centering
    \includegraphics[width=\linewidth]{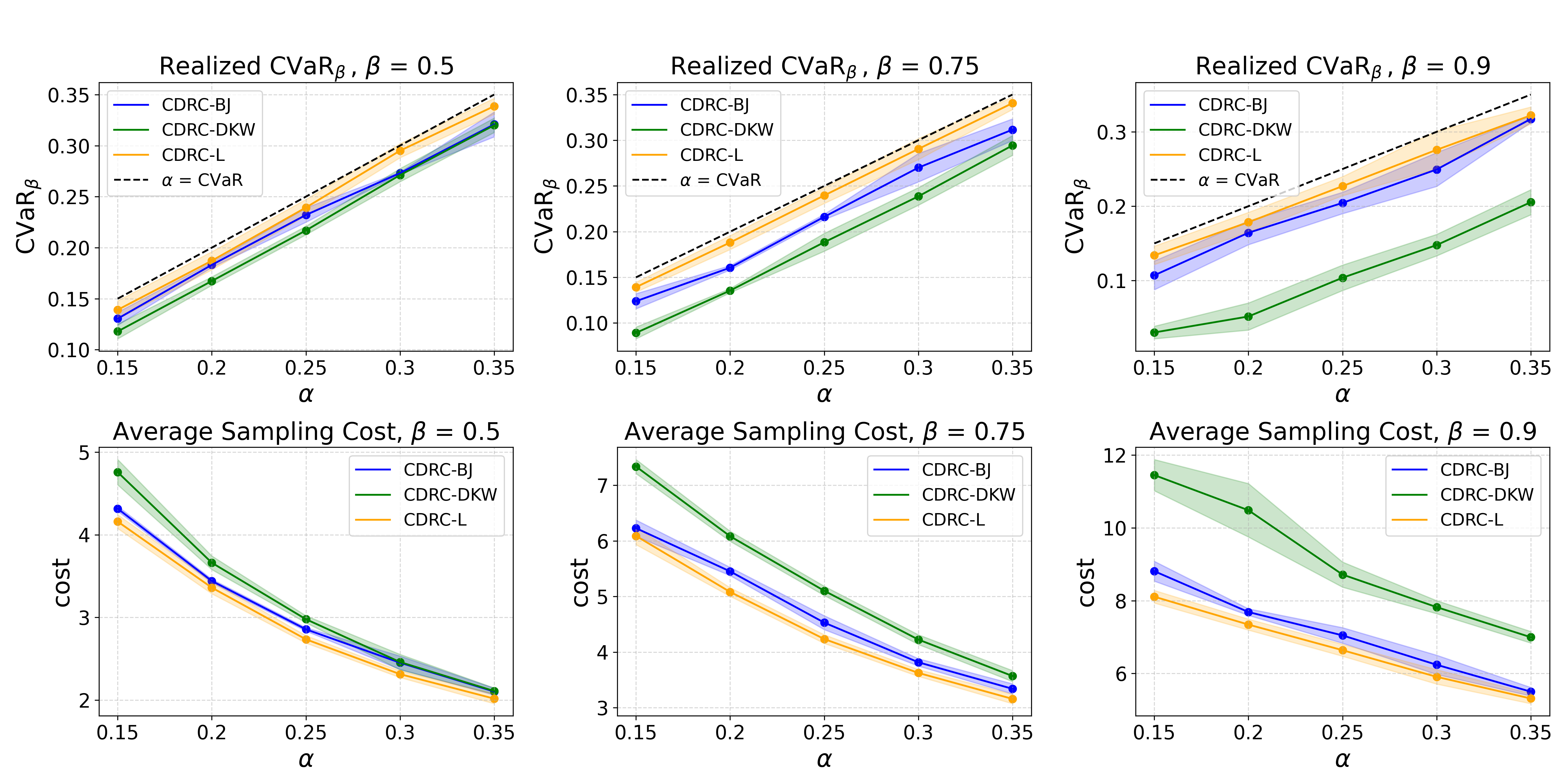}
    \caption{\textbf{Average sampling cost vs. $\alpha$ (row 1), and realized $\operatorname{CVaR}$ vs. $\alpha$ (row 2) for Spearman correlation between human and machine toxicity scores at $\rho = 0.57$ evaluated on held-out dataset}.  The confidence band is computed by taking the mean estimate plus/minus one standard error estimated from the results across independent experiments. Each subplot in the respective rows illustrates a different setting of $\beta \in \{0.5, 0.75, 0.9\}$. From the panels in the first row, we observe that our method, CDRC-L ({\color{orange} orange}), is an improvement to CDRC-DKW ({\color{green}green}) and CDRC-BJ ({\color{blue}blue}), as it is able to achieve risk control (shown in the black dotted line) while being less conservative than both baseline methods. Evident from the panels in the second row, our method is more efficient in generating a risk-controlled LLM response than DKW or BJ.}
    \label{fig:rho_57}
\end{figure*}

\begin{figure*}
    \centering
    \includegraphics[width=\linewidth]{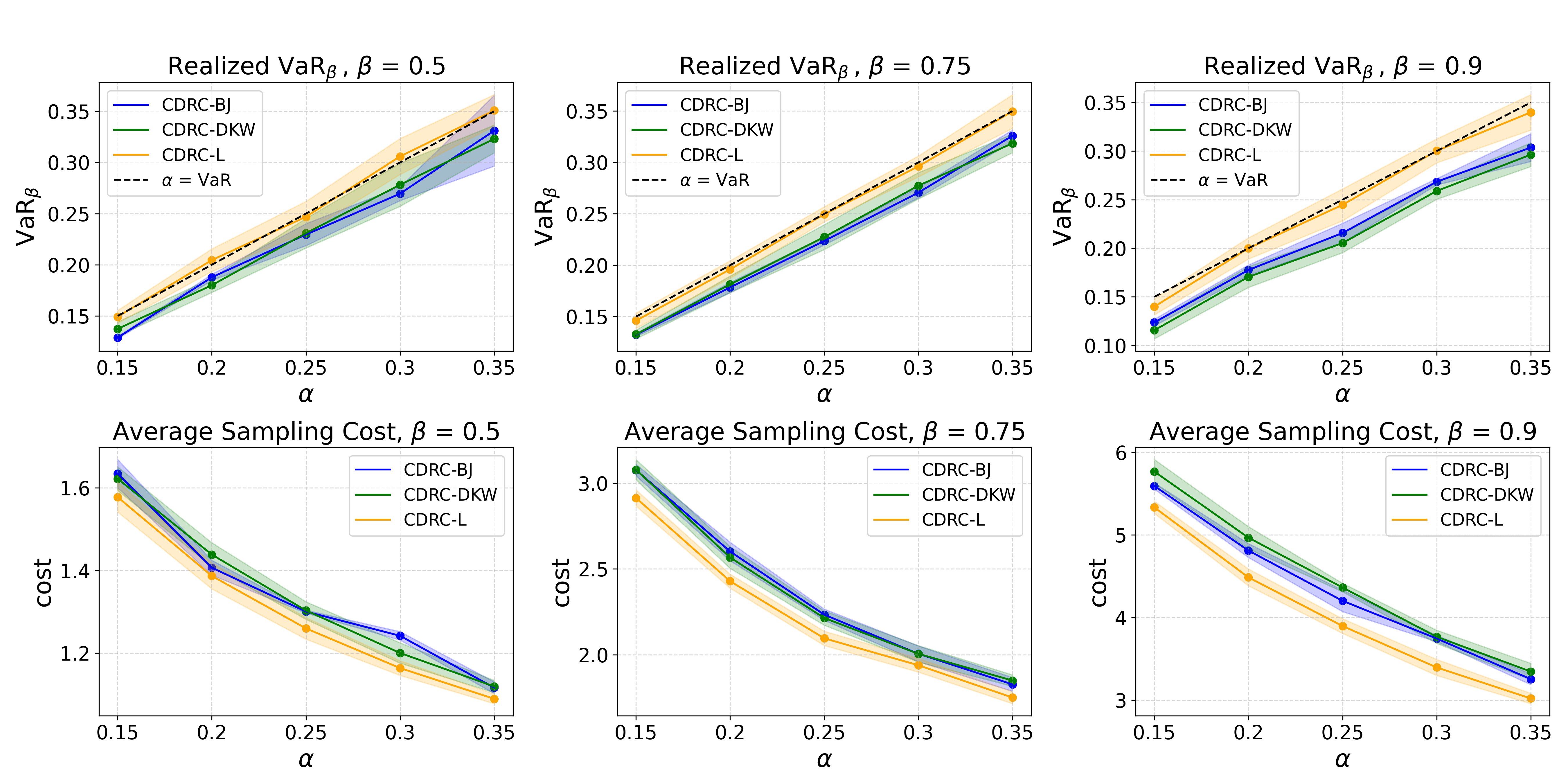}
    \caption{\textbf{Average sampling cost vs. $\alpha$ (row 1), and realized $\operatorname{VaR}$ vs. $\alpha$ (row 2) for Spearman correlation between human and machine toxicity scores at $\rho = 0.57$ evaluated on held-out dataset}. The confidence band is computed by taking the mean estimate plus/minus one standard error estimated from the results across the independent experiments. Each panel in the respective rows illustrates a different setting of $\beta \in \{0.5, 0.75, 0.9\}$. CDRC-L ({\color{orange} orange}), is an improvement to CDRC-DKW ({\color{green}green}) and CDRC-BJ ({\color{blue}blue}), as it is able to achieve risk control within the margin of error (shown in the black dotted line) while being more efficient than both baseline methods. }
    \label{fig:var_0.57}
\end{figure*}

\begin{figure*}
    \centering
    \includegraphics[width=0.95\linewidth]{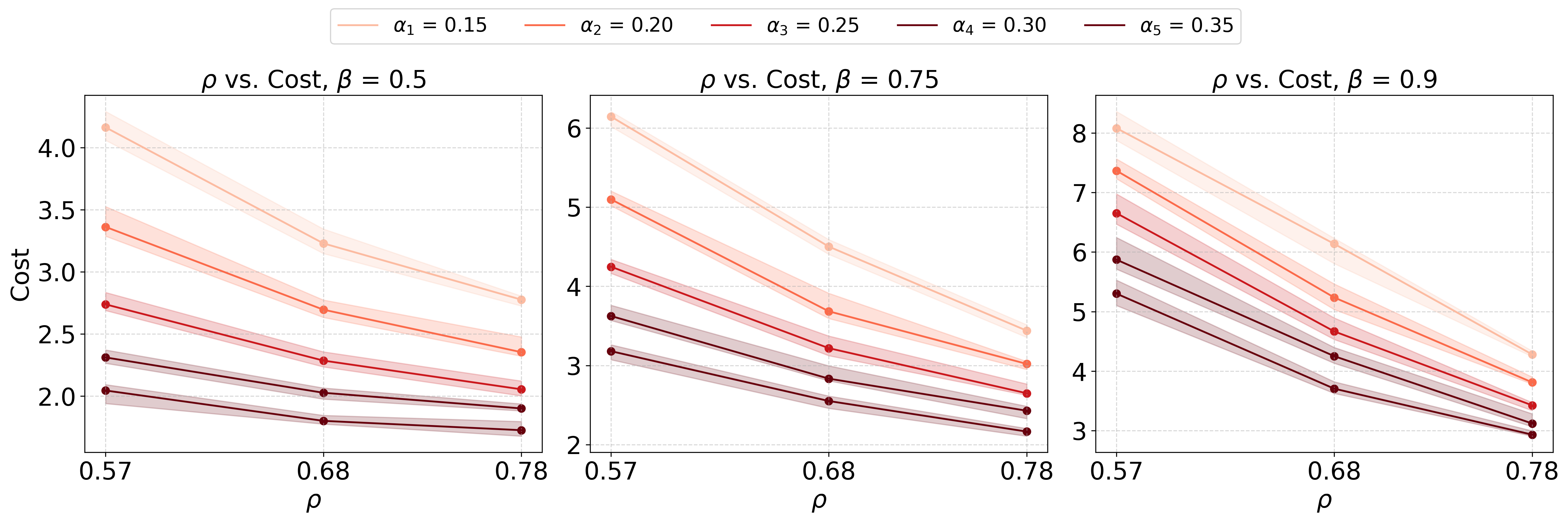}
    \caption{\textbf{Spearman correlation vs. average sampling cost for $\operatorname{CVaR}$}.}
    \label{fig:rho_cost}
\end{figure*}

Fig. \ref{fig:rho_57} shows the realized $\operatorname{CVaR}_\beta$ of human scores and the average sampling cost with $\rho = 0.57$ on the held-out dataset as functions of $\alpha$ for $\beta \in \{0.5, 0.75, 0.9\}$. {The panels in the first row shows that all methods control the risk at the target level and CDRC-L is least conservative, as discussed in Section \ref{subsec:DKW_BJ}. As a result, it incurs the smallest deployment cost among all three methods. Moreover, as $\beta$ increases, the advantage of CDRC-L is more prominent. Although CDRC-BJ improves upon CDRC-DKW due to the tighter bounds for extreme quantiles, it still underperforms CRDC-L. On the other hand, the gap between the target and realized $\operatorname{CVaR}_\beta$ stays nearly constant for CDRC-L across different values of $\beta$, suggesting that L-statistics are adaptive to different choices of $\psi$. }
 Results for $\rho = 0.68$, and $\rho = 0.78$ are presented in Appendix \ref{appendix:addtional_results_cvar}, showing similar patterns. 

{We present the results for $\operatorname{VaR}_\beta$ control with $\rho = 0.57$ in Fig. \ref{fig:var_0.57}. The comparison between CDRC-L, CDRC-DKW, and CDRC-BJ are qualitatively similar to $\operatorname{CVaR}_\beta$ control. Notably, the realized $\operatorname{VaR}_\beta$ of CDRC-L is extremely close to the target level, implying that it is not conservative in any way. Results for $\rho = 0.68$, and $\rho = 0.78$ are presented in Appendix \ref{sec:var}.}

Next, we examine how the deployment cost varies with the misalignment between human and machine ratings.  Fig. \ref{fig:rho_cost} demonstrates that, for all settings of $\beta$, as the Spearman correlation coefficient increases,  the cost of generating a $\operatorname{CVaR}$-controlled LLM response drops. This confirms our intuition that better-aligned machine ratings reduces the cost of calibration. Similar trends are observed for $\operatorname{VaR}$ control, as shown in Appendix \ref{sec:var}.

{
\section{Conclusion and discussion}

In this paper, we introduce an L-statistics-based calibration method that provably controls distortion risk measures for any black-box machine learning models. We showcase the potential of the method on LLM alignment, where conventional risk measures, such as the expectation of a loss function, fails to capture the tail risk. Our calibration approach is lightweight computationally and does not require retraining the model. We close the paper by discussing a few potential generalizations and extensions of our method.

~\\
\textbf{Recalibrating machine evaluation.} Machine-generated scores often do not have an interpretable scale. Ideally, the score should reflect the risk if a response with a score below it were accepted in the next task. Our method generates the cutoff $\hat{\lambda}(\alpha)$ for each target risk level $\alpha$. For each new LLM response with machine score $r$, we could invert $\hat{\lambda}(\alpha)$ to obtain a new score $\min\{\alpha: \hat{\lambda}(\alpha)\ge r\}$, which measures the smallest risk level at which the response is rejected. We will leave the theoretical investigation of this recalibrated score for future research. 

~\\
\noindent \textbf{Calibrating LLMs under context shifts. } Theorem \ref{thm:drc_rcps} only works for the population of prompts that has the same distribution as the data the method operates on. It does not necessarily hold when the data environment changes. In the presence of distribution shifts, we can potentially adapt the reweighting technique for traditional expected risk measures \cite{tibshirani2019conformal,lei2021conformal,candes2023conformalized} to correct for bias for L-statistics. 

~\\
\noindent \textbf{Uilizing preference data instead of direct human ratings. } Due to the rise of RLHF, preference data has become increasingly prevalent. While the above method only works with direct human ratings, there is a potential of adaptation to handle preference data, which exhibits a U-statistic structure. 
}
% \section*{Impact Statement}

% This paper presents work whose goal is to advance the field of 
% machine learning by ensuring safe deployment and alignment of large language models (LLMs). There are many potential societal consequences of our work, for instance, addressing the potential misalignment between human and machines, recalibrating machines such that they are better aligned with humans, and controlling the tail risks associated with language models. In this project, we did not use directly use data from humans. There are not negative impacts which we feel must be specifically highlighted here.

\bibliography{references}
\bibliographystyle{abbrv}

\newpage
\appendix
\onecolumn

\section*{Supplementary Materials}
In the supplementary materials, we provide implementation details, additional experimental results, and formalized proofs. Code repository can be found at
\url{https://anonymous.4open.science/r/distortion-risk-control/}.

\section{Implementation Details} \label{sec:appendix}

\subsection{Algorithm to generate the candidate set \texorpdfstring{$\mathcal{C}$}{C}}
\label{subsec:conformal_language_model}\label{sec:candidateresponse}
\begin{figure}[h!]
    \centering
    \includegraphics[width=\linewidth]{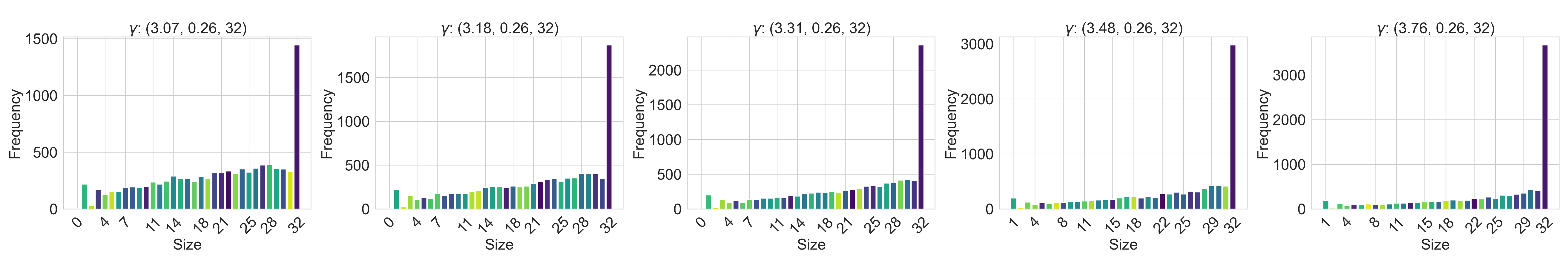}
    \caption{\textbf{Size of generated sets under different combinations of hyperparameters}. $\boldsymbol{\gamma} = (\gamma_1, \gamma_2, \gamma_3)$ refers to (preplexity, similarity, stopping threshold), respectively.}
    \label{fig:distribution}
\end{figure}

Following the original implementation of \textsc{LlaMA-2-7B-hf} model, we set
the generation temperature at 0.8 and the top-p parameter at 0.95. \label{sec:clm}
We generate candidate response sets for our DRC framework using Algorithm \ref{alg:conformal-sampling-with-rejection}, following \cite{quach2024conformallanguagemodeling}. To ensure quality and diversity of generated candidates, we filter responses by retaining only those with a Perpexity less than $2.61$ while ensuring that the \textsc{ROUGE-L} scores between samples in the candidate set is not greater than $0.26$. Table \ref{tab:clm} presents the percentiles of these metrics across all generated responses. Fig. \ref{fig:distribution} outlines the cardinality of the sets generated by various combinations of $\gamma$. 

\begin{algorithm}
    \caption{Generation of the candidate set $\mathcal{C}(x)$}
    \label{alg:conformal-sampling-with-rejection}
    \begin{algorithmic}
        \Require an input prompt $x$, a set-based confidence function $\mathcal{F}$, a text similarity function $\mathcal{S}$, a sample quality estimator $\mathcal{Q}$, a fixed threshold configuration $\boldsymbol{\lambda} = (\lambda_1, \lambda_2, \lambda_3)$, and a sampling budget $k_{\operatorname{max}}$, with conditional output $p_{\theta}(y \mid x)$ from a generative model. 
        \Function{CandidateSet}{$x, \mathcal{F}, \mathcal{S}, \mathcal{Q}, \boldsymbol{\lambda}, k_{\operatorname{max}}$}
        \State $\mathcal{C} \gets \{\}$
        \For{$k = 1, 2, \ldots, k_{\operatorname{max}}$}
            \State $y_k \gets y \sim p_{\theta}(y \mid x)$ \Comment{Sample from generative model}
            \If{$\mathcal{Q}(x, y_k) < \lambda_1$ and $\operatorname{max}\{\mathcal{S}(y_k, y_j): y_j \in \mathcal{C}\} > \lambda_2$} \Comment{Quality and similarity control}
                \State $\mathcal{C} \gets \mathcal{C} \cup \{y_k\}$
            \EndIf
            \If{$\mathcal{F}(\mathcal{C}) \geq \lambda_3$}    \Comment{Set-based confidence guarantee}
                \Break 
            \EndIf
        \EndFor
        \Return{$\mathcal{C}$}
        \EndFunction
    \end{algorithmic}
\end{algorithm}

\begin{table}[h!]
    \centering
    \caption{Percentiles of different hyperparameters.}
    \begin{tabular}{c c c c c c c}
    \toprule
        Percentile &  50 &  75 &  80 &  85 &  90 & 95 \\
        \midrule
         \textsc{ROUGE-L} &  0.26 &  0.39 &  0.43 &  0.47 &  0.52 & 0.59 \\
        Perplexity & 2.61 &  3.07 &  3.18 &  3.31 & 3.48 & 3.76 \\
        \bottomrule
    \end{tabular}
    \label{tab:clm}
\end{table}

\subsection{Detoxify model for \texorpdfstring{$r(\cdot)$}{r()} and \texorpdfstring{$r_m(\cdot)$}{rm()}}
We use the original Detoxify model as a proxy for human annotator scores, then finetune this base model with various sample sizes, a learning rate of $0.0001$, a batch size of $16$, and a weight decay of $3 \times 10^{-6}$ on a single Nvidia A40 GPU. The
Adam optimizer was employed with $\alpha=0.9$, $\beta=0.999$, and $\epsilon=10^{-8}$. We exclusively use the Detoxify framework to evaluate the text generated by our model without the prompts.
\subsubsection{A Semi-Synthetic Exercise Results} \label{sec:semisyn}
\begin{figure}
    \centering
    \includegraphics[width=0.5\linewidth]{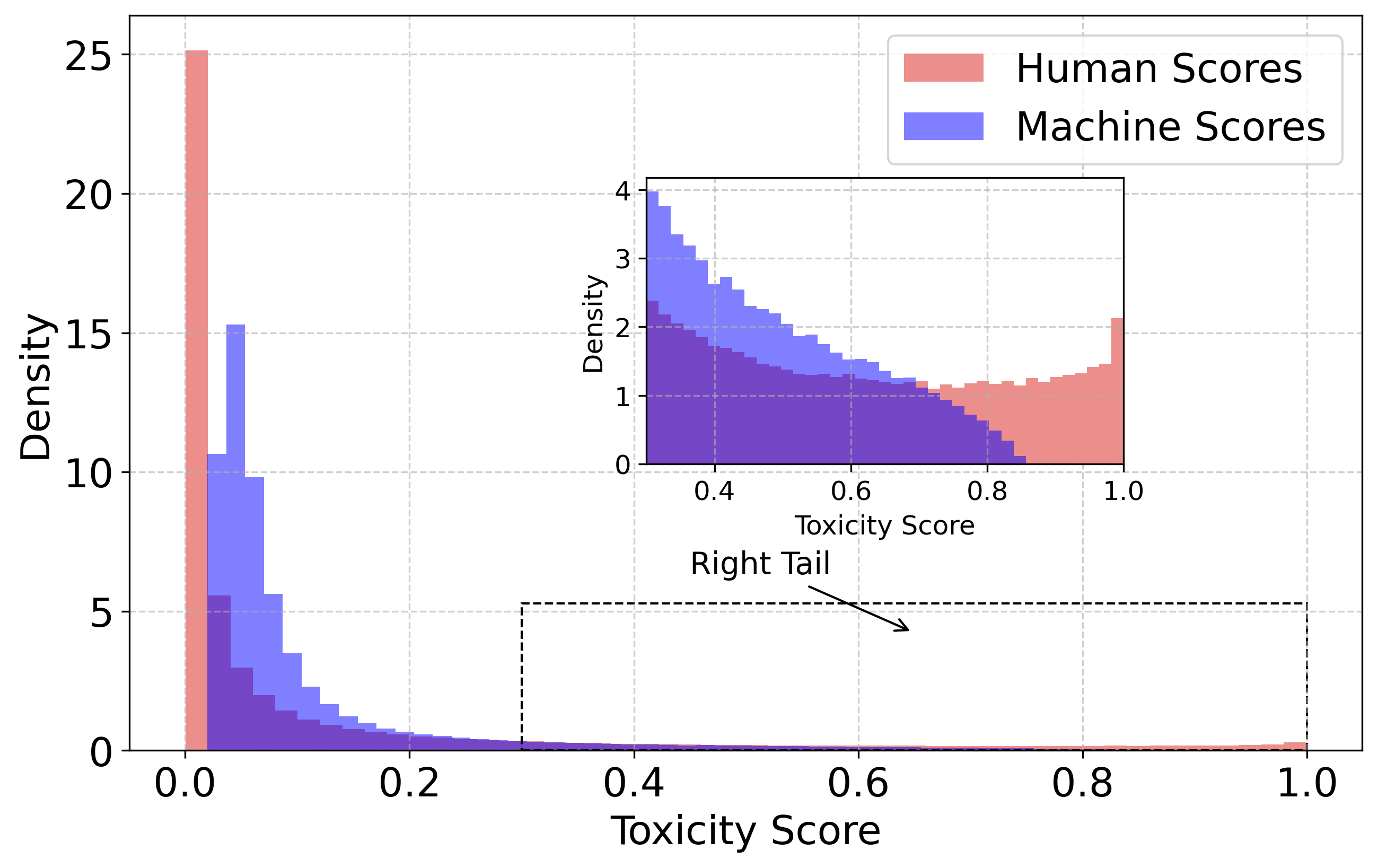}
    \caption{Distribution of toxicity scores assigned by humans (i.e., original Detoxify model) and machine (i.e., fine-tuned Detoxify model) for responses generated by the LLaMA 2 7B model across all prompts. The main plot shows the density of scores, highlighting a long-tailed distribution in both human and machine evaluations. The zoomed-in plot illustrates that human scores have a heavier tail than machine scores. }
    \label{fig:score_distribution}
\end{figure}

To study the effect of human-machine misalignment, we create a semi-synthetic data set of human and machine-generated toxicity scores. We use the Detoxify \cite{detoxify} model as our ``human" annotator, and a Detoxify model finetuned on a biased subsample of toxic instances as our ``machine" annotator. For details about implementation, see Section \ref{sec:toxicity}. Table \ref{tab:detoxify} outlines ROC-AUC (Area Under ROC Curve) comparison between the original Detoxify model, and the finetuned model, which illustrates that the original Detoxify model outperforms the finetuned model in predicting toxicity. Fig. \ref{fig:score_distribution} illustrates the distribution of human and machine-assigned scores for responses generated by the LLaMA 2 7B model across all sampled prompts. Both human and machine toxicity scores exhibit a long-tailed distribution in the real-world dataset. Due to the Detoxify model being fine-tuned with an emphasis on ``severe toxicity" samples, its scores are predominantly concentrated in the lower range, highlighting a misalignment between human and machine assessments.

\begin{table}[h!]
    \centering
    \caption{Performance comparison of detoxify models.}
    \begin{tabular}{c c}
        \toprule
        Model & ROC-AUC\\
        \midrule
        Detoxify (original) & 0.97 \\
        Finetuned Detoxify ($\rho=0.57$) & 0.86 \\
        \bottomrule
    \end{tabular}
    \label{tab:detoxify}
\end{table}

\subsection{Choices of \texorpdfstring{$\alpha$}{a}} \label{sec:alpha}
To select an appropriate value of $\alpha$, we arrange $r_{\lambda}(x)$ in increasing order then rule out the top $q\%$ responses, and compute the desired distortion risk of the lower $(1-q) \%$. The calculated value is used as the target $\alpha$. For details about implementation, see Section \ref{sec:methods-baselines}. Table \ref{tab:alpha} outlines the rational $\alpha$ values under different settings of $\lambda$ for different $q\%$ given that we are interested in studying $\operatorname{CVaR}$.
\begin{table}[h!]
    \centering
    \caption{Choice of $\alpha$ with different $q$\% under various settings of $\boldsymbol{\gamma}$ for $\operatorname{CVaR}$.}
    \begin{tabular}{c c c c c c}
        \toprule
         $\boldsymbol{\gamma} = (\gamma_1, \gamma_2, \gamma_3)$ &  $q = 1\%$  & $q = 5\%$ &  $q = 10\% $ & $q = 15\%$ & $q = 20\% $\\
        \midrule
        (3.07, 0.26, 32) & 0.815 & 0.580 & 0.356  & 0.217  & 0.145 \\
        (3.18, 0.26, 32) & 0.809 & 0.578 & 0.356  & 0.216  & 0.140 \\
        (3.31, 0.26, 32) & 0.815 & 0.574 & 0.354  & 0.219  & 0.142 \\
        (3.48, 0.26, 32) & 0.807 & 0.578 & 0.359  & 0.217  & 0.141 \\
        (3.76, 0.26, 32) & 0.815 & 0.589 & 0.352  & 0.221  & 0.141 \\
        \bottomrule
    \end{tabular}
    \label{tab:alpha}
\end{table}

\section{Additional experimental results for \texorpdfstring{$\operatorname{CVaR}$}{CVaR} control}

\subsection{Impact of Human-Machine Misalignment}
\label{appendix:addtional_results_cvar}
\begin{figure}[h!]
    \centering
    \includegraphics[width=\linewidth]{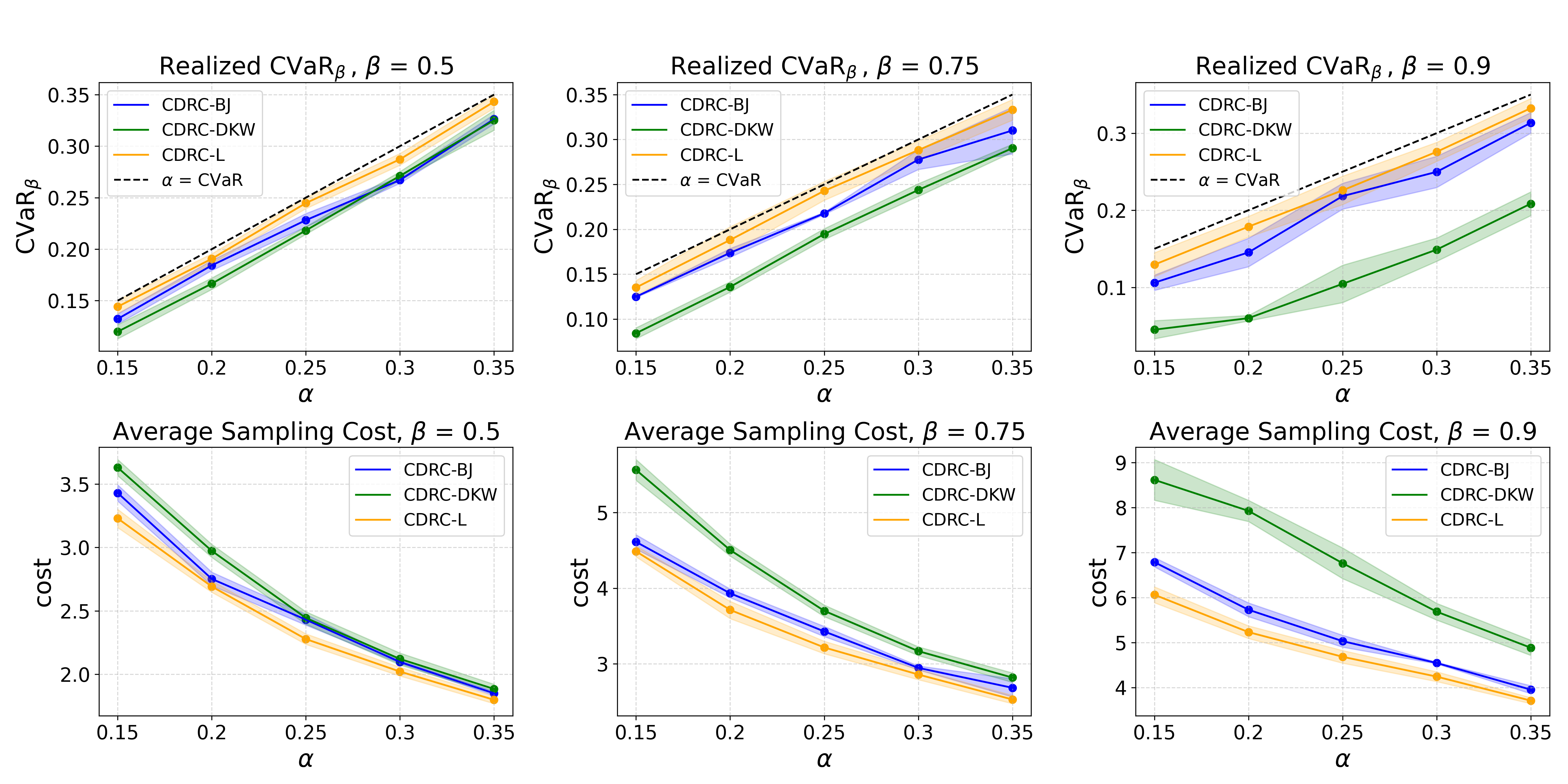}
    \caption{\textbf{Average cost and realized CVaR on held-out dataset for $\rho = 0.68$}.}
    \label{fig:rho_68}
\end{figure}

\begin{figure}[h!]
    \centering
    \includegraphics[width=\linewidth]{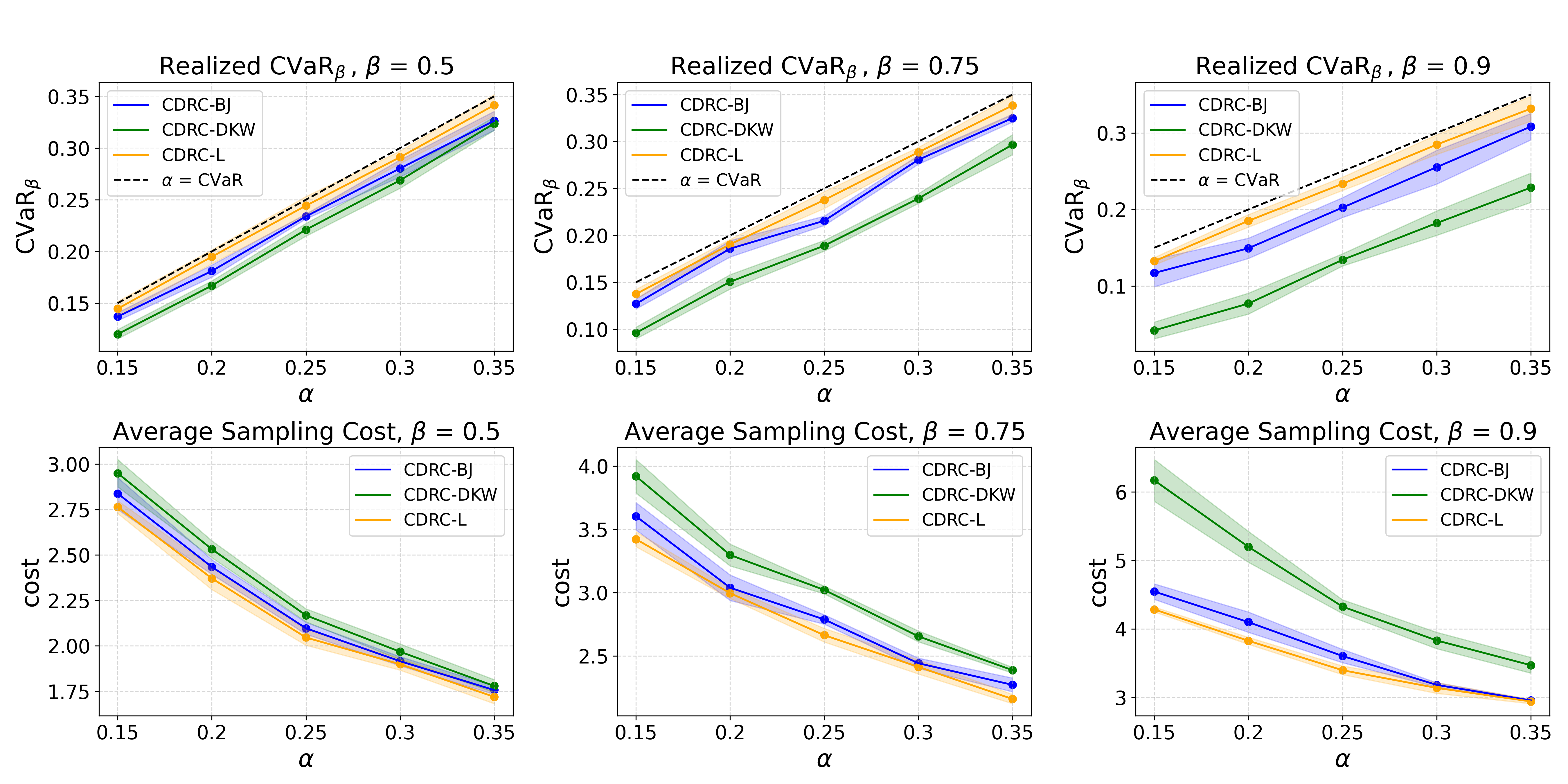}
    \caption{\textbf{Average cost and realized CVaR on held-out dataset for $\rho = 0.78$}.}
    \label{fig:rho_78}
\end{figure}

We evaluate the performance, i.e., average sampling cost, and realized CVaR on a held-out dataset, of our framework CDRC-L, against baseline methods CDRC-DKW and CDRC-BJ.  Fig. \ref{fig:rho_57}, Fig. \ref{fig:rho_68}, and Fig. \ref{fig:rho_78} illustrate the setting when the Spearman correlation ($\rho$) between the human and machine scores is 0.57, 0.68, and 0.78, respectively. From these figures, it is evident that all methods effectively control the risk at the specified level $\alpha$, given that they all fall below the black dotted line (where $\alpha = \operatorname{CVaR}$). Our method, CDRC-L, is consistently the least conservative of all methods across different values of $\beta$, and $\rho$. Conversely, baseline methods appear to grow increasingly conservative as $\beta$ increases. As a consequence, we see that our method is consistently the least costly across all settings of $\beta$, and $\rho$.

\section{Proofs} \label{sec:proofs}

\subsection{Proof of Theorem \ref{thm:asymp-norm}}
We decompose the proof into two parts. 

\begin{theorem}[Asymptotic normality of L-statistics] \label{thm:asymp-norm}
Let $f_\lambda, F_\lambda$ denote the PDF and CDF of the score $r_i$, respectively, and $\hat{F}_{n,\lambda}$ denote the empirical CDF of the scores $r_1, \ldots, r_n$. Further, let
$$R_{\psi}(\lambda) = \int_0^1 F_{\lambda}^{-1}(p)d\psi(p) \quad \text{and} \quad \hat{R}_{\psi}(\lambda) = \int \hat{F}^{-1}_{n, \lambda}(p)d\psi(p) ,$$
for any measure $\psi$ that may not have a density (e.g., $\psi$ can be a point mass at $p_0 \in (0, 1)$). Suppose 
    \begin{enumerate}
        \item $r_i \in [a, b]$ for some $-\infty < a < b < \infty$ almost surely, with $\inf_{r\in [a, b]}f_\lambda(r)> 0$, and $\sup_{r\in [a, b]}f_\lambda^\prime(r)< \infty$.
        \item $\int \frac{d \psi(t)}{f_\lambda(F_{\lambda}^{-1}(t))} < \infty$.
    \end{enumerate}

then $\sqrt{n}(\hat{R}_{\psi}(\lambda) - R_{\psi}(\lambda)) \stackrel{d}{\rightarrow} N(0, \sigma^2(\lambda))$ where 
\[\sigma^2(\lambda) = \int_0^1 \int_0^1 \frac{p \wedge p^\prime - p \cdot p^\prime}{f_\lambda(F_{\lambda}^{-1}(p))f_\lambda(F_{\lambda}^{-1}(p^\prime))} d\psi(p)d\psi(p^\prime) < \infty.\]
\begin{proof}
    By the Bahadur-Kiefer representation \cite{kiefer1970} (see also Theorem E of \cite{csorgHo1978strong}) of sample quantiles we can write 
    
    \begin{align*}
    \Delta_{n} &\stackrel{\Delta}{=} \sup_{p\in [0,1]}\lvert (F_{n, \lambda}^{-1}(p) - F_{\lambda}^{-1}(p))f_\lambda(F_{\lambda}^{-1}(p)) - (F_{n, \lambda}(F_{\lambda}^{-1}(p))-p)\rvert\\
    & = O_p \left(\frac{\log(n)^{1/2}(\log\log n)^{1/4}}{n^{3/4}} \right) = O_p \left(\frac{\log(n)}{n^{3/4}} \right),
    \end{align*}
    equivalently, 
    $$\left \lvert (\hat{F}_{n, \lambda}^{-1}(p) - F_{\lambda}^{-1}(p)) - \frac{F_{n, \lambda}(F_{\lambda}^{-1}(p))-p}{f_\lambda(F_{\lambda}^{-1}(p))} \right \rvert = \frac{\Delta_{n}}{f_\lambda(F_{\lambda}^{-1}(p))}.$$
    
    Given this, we have that
    \begin{align*}
        \hat{R}_{\psi}(\lambda) - R_{\psi}(\lambda) & = \int_0^1 (\hat{F}^{-1}_{n, \lambda}(p) - F_{\lambda}^{-1}(p))d\psi(p)\\
        & = \int_0^1 \frac{F_{n, \lambda}(F_{\lambda}^{-1}(p))-p}{f_\lambda(F_{\lambda}^{-1}(p))} d\psi(p) + \delta_n
    \end{align*}

    where 
    $$\lvert \delta_n\rvert \leq \Delta_{n} \cdot \int_0^1 \frac{1}{f_\lambda(F_{\lambda}^{-1}(p))} d\psi(p) =  O_p \left(\frac{\log(n)}{n^{3/4}} \right)$$
    holds by \textit{assumption 2}.
    Thus we have that 
    \begin{align*}
        \sqrt{n}(\hat{R}_{\psi}(\lambda) - R_{\psi}(\lambda)) & = \sqrt{n} \cdot \frac{1}{n} \sum_{i=1}^n \int_0^1 \frac{\mathbbm{1}\{r_i \leq F_{\lambda}^{-1}(p)\}-p}{f_\lambda(F_{\lambda}^{-1}(p))} d\psi(p) + o_p(1) \\
        & \stackrel{\Delta}{=} \frac{1}{\sqrt{n}} \sum_{i=1}^n S(r_i) + o_p(1)
    \end{align*}
    By \textit{assumption 2},
    $$S(r_i) = \int_0^1 \frac{\mathbbm{1}\{r_i \leq F_{\lambda}^{-1}(p)\}-p}{f_\lambda(F_{\lambda}^{-1}(p))} d\psi(p) \leq \int_0^1\frac{1}{f_\lambda(F_{\lambda}^{-1}(p))}d\psi(p) < \infty.$$
    It remains to prove $\mathbb{E}[S(r_i)] = 0$ and $\mathrm{Var}(S(r_i)) = \mathbb{E}[S^2(r_i)] < \infty$. 

    To study the mean, note that by \textit{assumption 1}, $F_{\lambda}^{-1}(p)$ is uniquely defined and $\mathbb{P}(r_i \le F_\lambda^{-1}(p)) = p$. Thus, 
    $$\mathbb{E}[S(r_i)] = \int_0^1\frac{\mathbb{E}[\mathbbm{1}\{r_i \leq F_{\lambda}^{-1}(p)\}-p]}{f_\lambda(F_{\lambda}^{-1}(p))} d\psi(p)=0$$

    To study the variance, observe that the second moment can be expressed as follows
    \begin{align*}
        \mathbb{E}[S^2(r_i)] & = \mathbb{E} \left [\left ( \int_0^1\frac{\mathbbm{1}\{r_i \leq F_{\lambda}^{-1}(p)\}-p}{f_\lambda(F_{\lambda}^{-1}(p))} d\psi(p) \right ) ^2\right ] \\
        & = \mathbb{E} \left [\int_0^1\frac{\mathbbm{1}\{r_i \leq F_{\lambda}^{-1}(p)\}-p}{f_\lambda(F_{\lambda}^{-1}(p))} d\psi(p) \cdot \int_0^1\frac{\mathbbm{1}\{r_i \leq F_{\lambda}^{-1}(p^\prime)\}-p^\prime}{f_\lambda(F_{\lambda}^{-1}(p^\prime))} d\psi(p^\prime)\right ]\\
        & = \int_0^1 \int_0^1 \frac{\mathbb{E} \left [(\mathbbm{1}\{r_i \leq F_{\lambda}^{-1}(p)\}-p)\cdot (\mathbbm{1}\{r_i \leq F_{\lambda}^{-1}(p^\prime)\}-p^\prime)\right ]}{f_\lambda(F_{\lambda}^{-1}(p))f_\lambda(F_{\lambda}^{-1}(p^\prime))} d\psi(p)d\psi(p^\prime)
    \end{align*}

    Let $B(p) = \mathbbm{1}\{r_i \leq F_{\lambda}^{-1}(p)\}$. Notice that $\mathbb{E}[B(p)] = p$, thus
    \begin{align*}
        \mathbb{E} \left [(B(p)-p)\cdot (B(p^\prime)-p^\prime)\right ] & = \operatorname{Cov}(B(p), B(p^\prime))\\
        & = \mathbb{E} \left [B(p)\cdot B(p^\prime)\right ] - p \cdot p^\prime \\
        & = \begin{cases}
            p - p \cdot p^\prime \quad, \quad \text{if }p \leq p^{\prime}\\
            p^\prime - p \cdot p^\prime \quad, \quad \text{if }p > p^{\prime}\\
        \end{cases} \\
        & = p \wedge p^\prime - p \cdot p^\prime
    \end{align*}

    Therefore,
    $$\operatorname{Var}(S(r_i)) = \int_0^1 \int_0^1 \frac{p \wedge p^\prime - p \cdot p^\prime}{f_\lambda(F_{\lambda}^{-1}(p))f_\lambda(F_{\lambda}^{-1}(p^\prime))} d\psi(p)d\psi(p^\prime)$$
   
    By \textit{assumption 2}, 
    $$\operatorname{Var}(S(r_i))\leq \int_0^1 \int_0^1 \frac{1}{f_\lambda(F_{\lambda}^{-1}(p))f_\lambda(F_{\lambda}^{-1}(p^\prime))} d\psi(p)d\psi(p^\prime) = \left \lvert \int_0^1 \frac{1}{f_\lambda(F_{\lambda}^{-1}(p))} d\psi(p)\right \rvert ^2 < \infty.$$
\end{proof}
\end{theorem}

\begin{theorem}[Consistent variance estimate for L-Statistics]
    In the same setting as Theorem \ref{thm:asymp-norm}, assume the \textit{assumption 1} holds and that $\psi(y) = \int_{0}^{y}\psi^{\prime}(z) dz$ for some $\psi'$ that is bounded and continuous at $F_{\lambda}(r)$ for Lebesgue almost-every $r$. 
    
    Let $$\hat{\sigma}_n^2(\lambda) = \int_a^b\int_a^b \psi^{\prime}(\hat{{F}}_{n,\, \lambda}(r)) \psi^{\prime}(\hat{{F}}_{n,\, \lambda}(\tilde{r}))(\hat{{F}}_{n,\, \lambda}(r \wedge \tilde{r}) - \hat{{F}}_{n,\, \lambda}(r)\hat{{F}}_{n,\, \lambda}(\tilde{r}))\, dr \, d\tilde{r}.$$
    If $\psi^\prime$ is bounded,
    then $\hat{\sigma}_n^2(\lambda) \stackrel{{a.s.}}{\longrightarrow} \sigma^2(\lambda)$ as $n \rightarrow \infty.$
    
\begin{proof}
    We first verify \textit{assumption 2}. 
    Take $p = F_{\lambda}(x)$,
        \begin{align*}
            \int_0^1 \frac{1}{f_\lambda(F_{\lambda}^{-1}(p))} d\psi(p) & = \int \frac{\psi^\prime(p)}{f_\lambda(F_{\lambda}^{-1}(p))} dp \\
            & = \int_{a}^b \frac{\psi^\prime(F_{\lambda}(x))}{f_\lambda(x)} dF_{\lambda}(x) \\
            & = \int_{a}^b \psi^\prime(F_{\lambda}(x)) dx.
        \end{align*}
        The integral is bounded since $\psi'$ is bounded and $a, b$ are both finite.
    Let $p = F_{\lambda}(r)$, and $p^\prime = F_{\lambda}(\tilde{r})$ in the double integral of $\sigma^2(\lambda)$, then 
    \begin{align*}
        \sigma^2(\lambda) & = {\int_0^1 \int_0^1 \frac{p \wedge p^\prime - p \cdot p^\prime}{f_\lambda(F_{\lambda}^{-1}(p))f_\lambda(F_{\lambda}^{-1}(p^\prime))} d\psi(p)d\psi(p^\prime)}\nonumber \\
        & = \int_{a}^b \int_a^b \frac{F_{\lambda}(r \wedge \tilde{r})- F_{\lambda}(r)F_{\lambda}(\tilde{r})}{\cancel{f_\lambda(r)f_\lambda(\tilde{r})}} \psi^\prime(F_{\lambda}(r))\psi^\prime(F_{\lambda}(\tilde{r}))\cdot \cancel{f_\lambda(r)f_\lambda(\tilde{r})}\, dr\, d\tilde{r}. 
    \end{align*}\label{eq:var_alternative_formula}
    
    By the Glivenko-Centelli Theorem, 
    $\hat{F}_{n, \lambda}(r) \stackrel{a.s.}{\longrightarrow} F_\lambda(r)$ for every $r\in [a, b]$, thus, for any $r, r'\in [a, b]$, 
    $$\hat{{F}}_{n,\, \lambda}(r \wedge \tilde{r}) - \hat{{F}}_{n,\, \lambda}(r)\hat{{F}}_{n,\, \lambda}(\tilde{r}) \stackrel{a.s.}{\longrightarrow} F_{\lambda}(r \wedge \tilde{r})- F_{\lambda}(r)F_{\lambda}(\tilde{r}).$$
    Furthermore, since $\psi'$ is continuous at $F(r)$ for almost every $r$ under the Lebesgue measure, by the continuous mapping theorem, 
    $$\psi^\prime (\hat{{F}}_{n,\, \lambda}(r)) \stackrel{a.s.}{\longrightarrow} \psi^\prime (F_{\lambda}(r))$$
    for almost every $r$ under Lebesque measure. 
    Suppose that $\psi^\prime$ is bounded above by $B$, then
    $$\psi^{\prime}(\hat{{F}}_{n,\, \lambda}(r)) \psi^{\prime}(\hat{{F}}_{n,\, \lambda}(\tilde{r}))(\hat{{F}}_{n,\, \lambda}(r \wedge \tilde{r}) - \hat{{F}}_{n,\, \lambda}(r)\hat{{F}}_{n,\, \lambda}(\tilde{r})) \leq B^2,$$
    thus
    $$\hat{\sigma}_n^2(\lambda) \stackrel{{a.s.}}{\longrightarrow} \sigma^2(\lambda)$$
    by the Dominated Convergence Theorem. 
    \end{proof}
\end{theorem}

\subsection{Proof of Corollary \ref{cor:CVaR_var}}

For $\mathrm{CVaR}_\beta$, $\psi(p) = \max\{p - \beta, 0\} / (1 - \beta)$ and hence $\psi' = I(p\ge \beta) / (1 - \beta)$. By \eqref{eq:var_alternative_formula},
\begin{align*}
\sigma^2(\lambda) &= \frac{1}{(1 - \beta)^2}\int_{a}^b\int_{a}^{b} (F_{\lambda}(r \wedge \tilde{r})- F_{\lambda}(r)F_{\lambda}(\tilde{r}))I(F_\lambda(r) \ge \beta) I(F_\lambda(\tilde{r})\ge \beta) dr d\tilde{r}\\
& = \frac{1}{(1 - \beta)^2}\int_{F_\lambda^{-1}(\beta)}^b\int_{F_\lambda^{-1}(\beta)}^b (F_{\lambda}(r \wedge \tilde{r})- F_{\lambda}(r)F_{\lambda}(\tilde{r})) dr d\tilde{r}.
\end{align*}
Let $G_\lambda$ be the CDF of $r'_i \triangleq \max\{r_i, \beta\}$. Then 
\[G_\lambda(r) = F_\lambda(r) I(r \ge F_\lambda^{-1}(\beta)).\]
By Hoeffding's covariance identity \citep{hoeffding1940masstabinvariante}, 
\[\mathrm{Var}(r'_i) = \int_a^b \int_a^b (G_{\lambda}(r \wedge \tilde{r})- G_{\lambda}(r)G_{\lambda}(\tilde{r})) dr d\tilde{r} = \int_{F_\lambda^{-1}(\beta)}^b\int_{F_\lambda^{-1}(\beta)}^b (F_{\lambda}(r \wedge \tilde{r})- F_{\lambda}(r)F_{\lambda}(\tilde{r})) dr d\tilde{r}.\]
Thus, 
\[\sigma^2(\lambda) = \frac{1}{( 1- \beta)^2}\mathrm{Var}(r'_i).\]
Since $r'_i$ is bounded, the Law of Large number implies that $\mathrm{Var}(r'_i)$ can be estimated consistently by the empirical variance of $(r'_1, \ldots, r'_n)$.

\subsection{Proof of Theorem \ref{thm:drc_rcps}}
The proof is very similar to \cite{dfrcps}, except that they only consider expected risk measures. If $R_\psi(\lambda)\le \alpha$ for all $\lambda\in \Lambda$, then the result obviously holds. Assume $\sup_{\lambda\in \Lambda}R_\psi(\lambda) > \alpha$. Since $R_\psi(\lambda)$ is continuous and strictly increasing, it crosses $\alpha$ exactly once. Let $\lambda^*$ denote the crossing point, i.e., $R_\psi(\lambda^*) = \alpha$. Then 
\[R_\psi(\hat{\lambda}) > \alpha \Longleftrightarrow \hat{\lambda} > \lambda^*\Longrightarrow \hat{R}_\psi^+(\lambda^*)\le \alpha,\]
where the last line is due to the definition of $\hat{\lambda}$. By Theorem \ref{thm:drc}, 
$$\lim_{n\rightarrow \infty}\mathbb{P}(\hat{R}_\psi^+(\lambda^*) < R_\psi(\lambda^*)) = \delta.$$
Since $R_\psi(\lambda^*) = \alpha$, we have
\[\limsup_{n\rightarrow \infty}\mathbb{P}(R_\psi(\hat{\lambda}) > \alpha)\ge \limsup_{n\rightarrow \infty}\mathbb{P}(\hat{R}_\psi^+(\lambda^*) < R_\psi(\lambda^*)) = \delta.\]
The proof is then completed. 

\begin{figure}[h!]
    \centering
    \includegraphics[width=\linewidth]{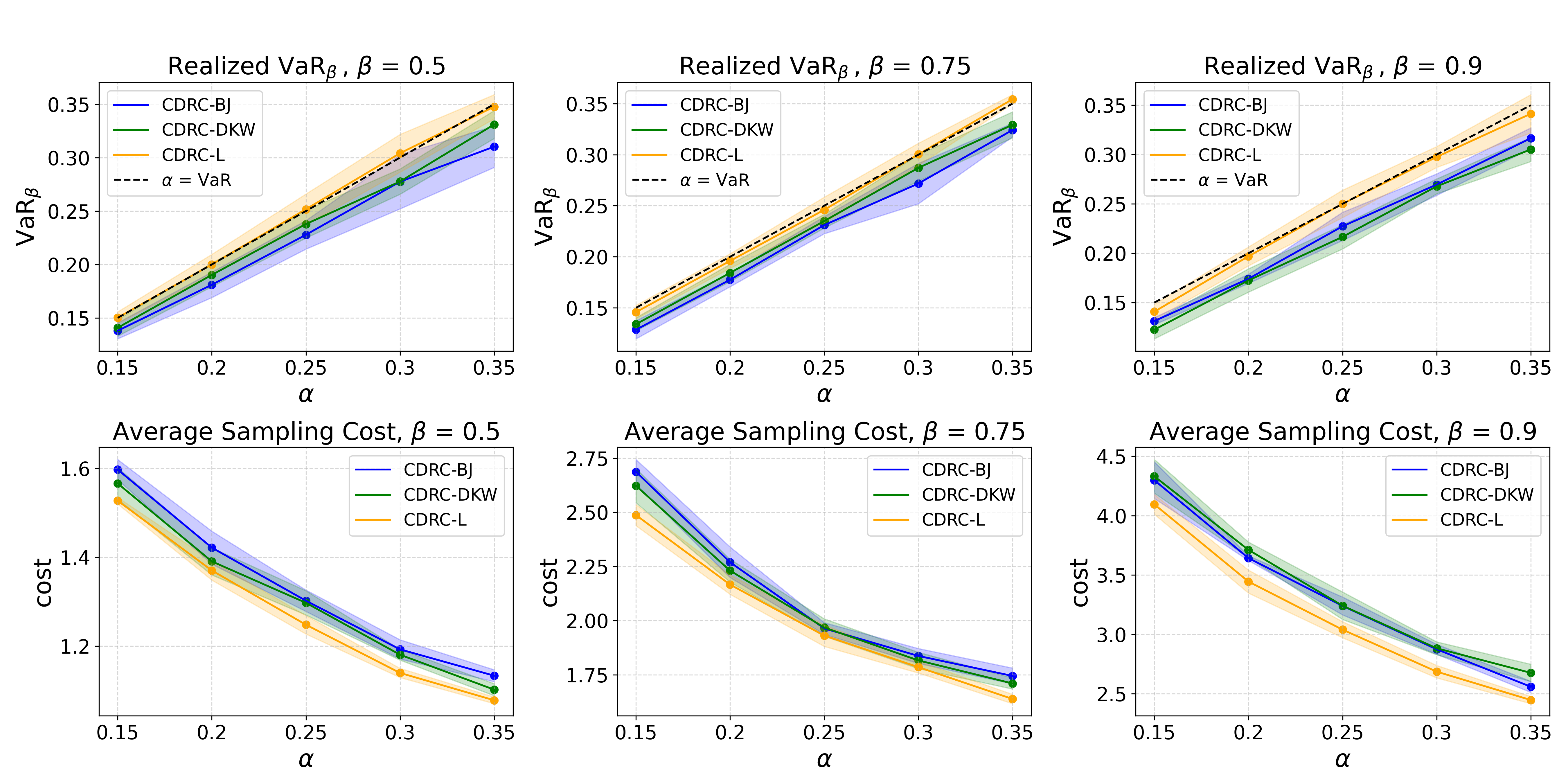}
    \caption{\textbf{Average cost and realized $\operatorname{VaR}$ on held-out dataset for $\rho = 0.68$}.}
    \label{fig:var_0.68}
\end{figure}

\section{Distortion Risk Control for {VaR}} \label{sec:var}

\subsection{Asymptotics of empirical quantiles}
Take $\psi(q) = \delta_\beta(q)$, we obtain $\operatorname{VaR}_\beta(\lambda) = F^{-1}_\lambda(\beta)$.  Theorem \ref{thm:drc} implies that the empirical $\beta$-th quantile 
 is asymptotically normal with variance
$$\sigma^2_{\operatorname{VaR}_\beta}(\lambda) = \frac{\beta(1 - \beta)}{f_\lambda^2(F_\lambda^{-1}(\beta))}.$$
Empirically, we use the standard Bootstrap procedure to estimate $\sigma^2_{\operatorname{VaR}_\beta}(\lambda)$ with $1000$ samples.

\subsection{Additional experimental results for VaR} 
\begin{figure}[h!]
    \centering
    \includegraphics[width=\linewidth]{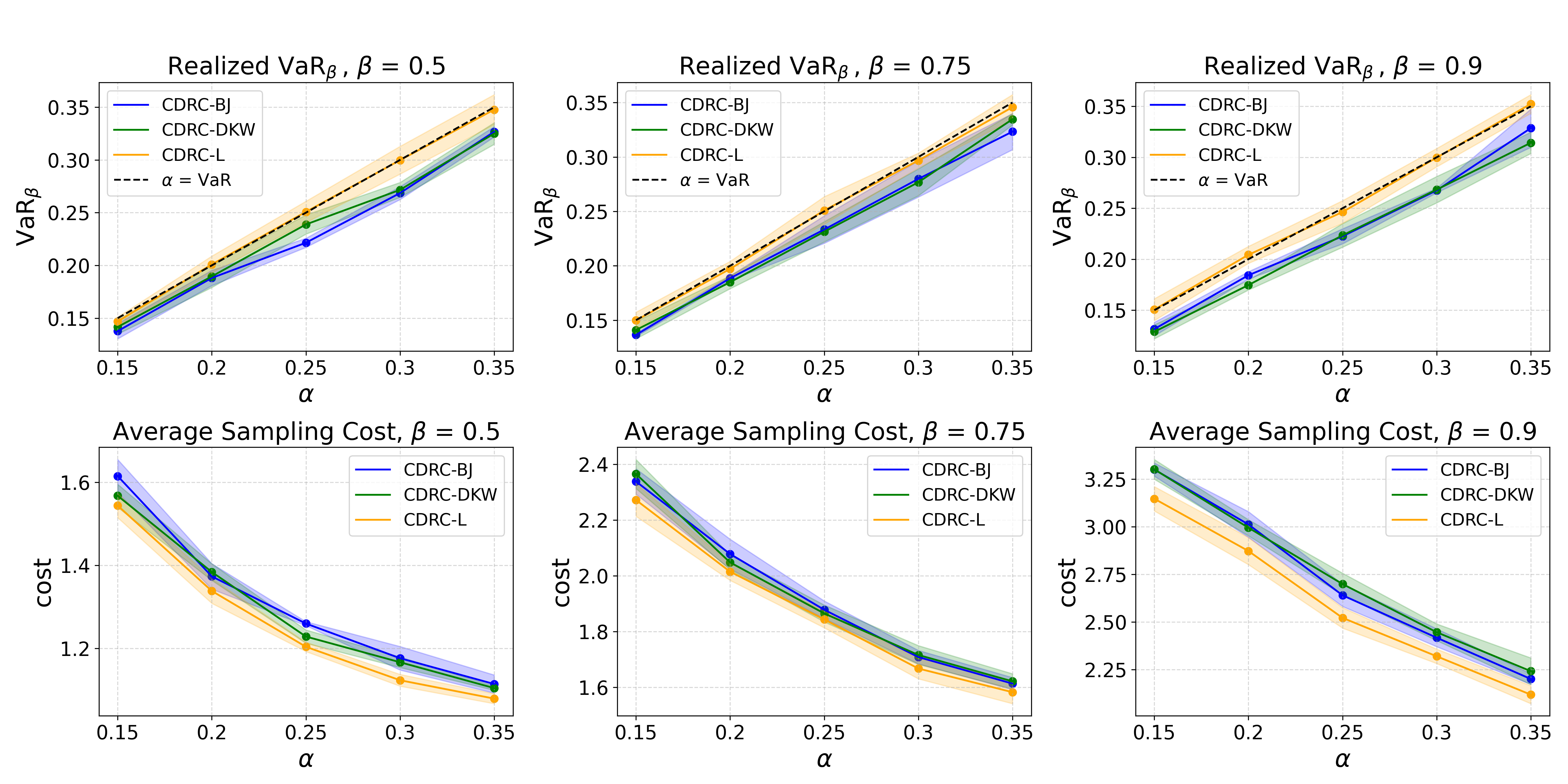}
    \caption{\textbf{Average cost and realized $\operatorname{VaR}$ on held-out dataset for $\rho = 0.78$}.}
    \label{fig:var_0.78}
\end{figure}

\begin{figure}[h!]
    \centering
    \includegraphics[width=\linewidth]{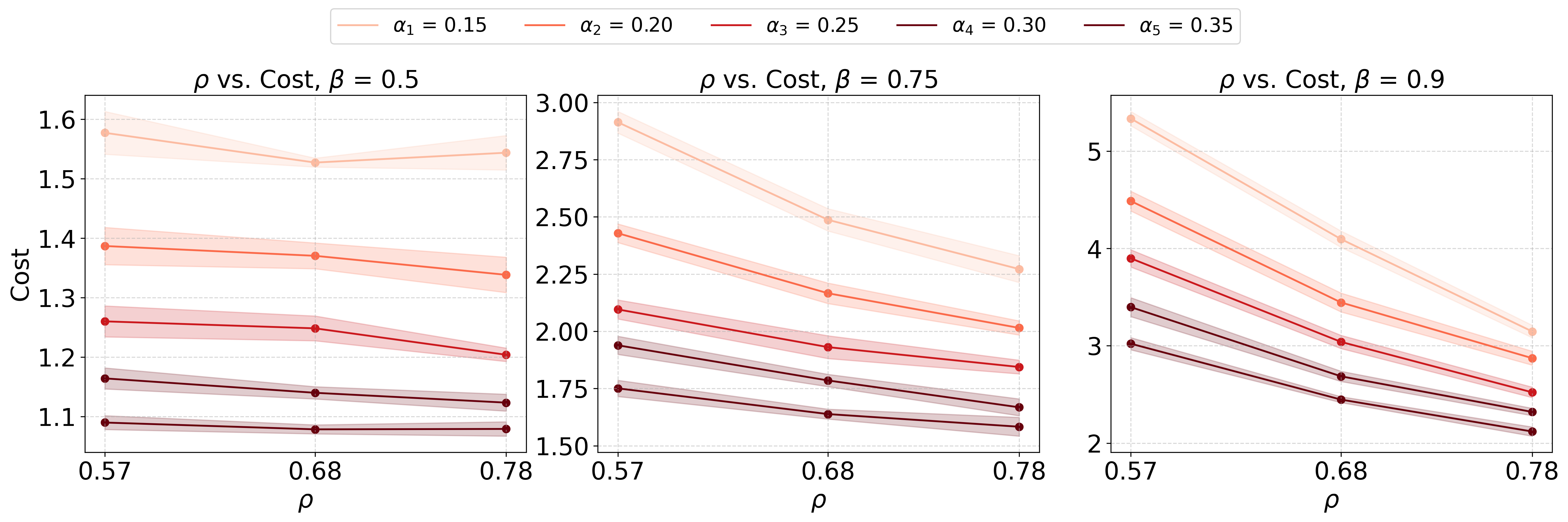}
    \caption{\textbf{Spearman correlation vs. average sampling cost for $\operatorname{VaR}$}.}
    \label{fig:var_rho}
\end{figure}

We evaluate the performance, i.e., average sampling cost, and realized VaR on a held-out dataset, of our framework CDRC-L, against baseline methods CDRC-DKW and CDRC-BJ. See Fig. \ref{fig:var_0.68} for $\rho = 0.68$, and Fig. \ref{fig:var_0.78} for $\rho = 0.78$. Similar to what we observe with CVaR, it is evident from the plots that all methods control the risk at the specified level $\alpha$ since they fall below the black dotted line (where $\alpha = \operatorname{VaR}$) within the margin of error. Again, our method, CDRC-L, is consistently the least conservative of all methods across different values of $\beta$, and $\rho$. As a result, we see that our method is consistently the least costly across all settings of $\beta$, and $\rho$. Fig. \ref{fig:var_rho} illustrates that for all settings of $\beta$, as the Spearman correlation coefficient increases,  the cost of generating a $\operatorname{VaR}$-controlled LLM response decreases.

\end{document}